\documentclass[11pt,oneside,final]{ntu_qe}

\newcommand{\numcol}{1}
\usepackage[svgnames, username, dvipsnames]{xcolor}
\usepackage{subfigure}
\usepackage{caption}
\usepackage{NTUdis}
\usepackage{setspace}
\usepackage{graphicx} 
\graphicspath{{Figures/}}
\usepackage{float}
\usepackage{algorithm}
\usepackage{amsfonts}
\usepackage{ifthen}
\usepackage{booktabs}
\usepackage{amssymb}
\usepackage{supertabular}
\usepackage{array}
\usepackage{multirow}
\usepackage{fancyhdr}
\usepackage{fancyheadings}
\usepackage{psfrag}
\usepackage{amsmath}
\usepackage{hhline}
\usepackage{type1cm}
\usepackage{lettrine}
\usepackage{cite}
\usepackage{fancybox}
\usepackage{ctable}
\usepackage{comment}
\usepackage{hyperref}
\usepackage[titletoc]{appendix}
\usepackage{pdflscape}
\usepackage{tabularx}
\usepackage{url}
\usepackage{ulem}
\usepackage{tabularx}
\usepackage{array}
\usepackage{colortbl}
\newcolumntype{Y}{>{\raggedleft\arraybackslash}X}
\newif\ifblackandwhite

\usepackage[hmargin=2cm,vmargin=2.5cm]{geometry}
\usepackage{etoolbox}
\usepackage{longtable}%

\usepackage{colortbl}%
\newcommand{\myrowcolour}{\rowcolor[gray]{0.925}}
\usepackage{booktabs}

\ifblackandwhite
  
  \newcommand{\highest}[1]{\textbf{#1}}
\else
  
  \newcommand{\highest}[1]{\textcolor{Maroon}{\textbf{#1}}}%
\fi

\makeatletter
\let\@currsize\normalsize
\makeatother

\newcommand{\checkOneCol}{\ifthenelse{\numcol = 1}}

\newcommand{\wabstract} {\begin{abstract}}
\newcommand{\wendabstract} {\end{abstract}}

\newcommand{\weqnarray}{\begin{eqnarray}}
\newcommand{\wendeqnarray} {\end{eqnarray}}
\newcommand{\waligneqnarray}{\begin{align}}
\newcommand{\walignendeqnarray} {\end{align}}

\newcommand{\wtable}{\begin{table}}
\newcommand{\wendtable}{\end{table}}

\newcommand{\wfigure}{\begin{figure}}
\newcommand{\wendfigure}{\end{figure}}

\newcommand{\wlist}{\begin{itemize}}
\newcommand{\wendlist}{\end{itemize}}
\newcommand{\wdesc}{\begin{description}}
\newcommand{\wenddesc}{\end{description}}

\newcommand{\wnumlist}{\begin{enumerate}}
\newcommand{\wendnumlist}{\end{enumerate}}

\newtheorem{theorem}{Theorem}
\newtheorem{lemma}{Lemma}
\newtheorem{corollary}{Corollary}
\newtheorem{definition}{Definition}
\newtheorem{example} {Example}
\newtheorem{case} {Case}
\newcommand{\wtheorem}{\begin{theorem}}
\newcommand{\wendtheorem}{\end{theorem}}
\newcommand{\wlemma}{\begin{lemma}}
\newcommand{\wendlemma}{\end{lemma}}
\newcommand{\wcorollary}{\begin{corollary}}
\newcommand{\wendcorollary}{\end{corollary}}
\newcommand{\wdefinition}{\begin{definition}}
\newcommand{\wenddefinition}{\end{definition}}
\newcommand{\wexample}{\begin{example}}
\newcommand{\wendexample}{\end{example}}
\newcommand{\wproof}{\begin{proof}}
\newcommand{\wendproof}{\end{proof}}
\newcommand{\wcase}{\begin{case}}
\newcommand{\wendcase}{\end{case}}
\newcommand{\ie}{i.e., }
\newcommand{\eg}{e.g., }
\newcounter{ifDRR}
\newcommand{\spellDRR}{Deficit Round Robin}
\newcommand{\abbrevDRR}{DRR}
\newcommand{\DRR}{%
\ifthenelse{\value{ifDRR}= 0}%
{\spellDRR\  (\abbrevDRR) \cite{scheduler:drr}\setcounter{ifDRR}{1}}%
{\abbrevDRR}%
}
\newcounter{ifGOODPUT}
\newcommand{\spellGOODPUT}{goodput}
\newcommand{\goodput}{%
\ifthenelse{\value{ifGOODPUT}=0}%
{\spellGOODPUT\  
\footnote{Goodput is the effective throughput, 
as determined by the data successfully
received and decoded at the receiver. }
\setcounter{ifGOODPUT}{1}}%
{\spellGOODPUT}%
}
\newcommand{\setonehalfspace}{\hsp}
\newcommand{\setdoublespace}{\dsp}

\begin{document}
\setdoublespace

\newcommand{\HRule}[1]{\rule{\linewidth}{#1}} 	

\thispagestyle{empty}

\makeatletter							
\def\printtitle{%
    {\centering \@title\par}}
\makeatother									

\makeatletter							
\def\printauthor{%
    {\centering \large \@author}}				
\makeatother							

\title{	\normalsize \textsc{} 	
		 	\\[2.0cm]								
			\HRule{0.5pt} \\						
			\LARGE \textbf{{Review of Video Predictive Understanding: Early Action Recognition and Future Action Prediction}}	
			\HRule{2pt} \\ [0.5cm]		
			\normalsize 
		}

\author{
		He Zhao and Richard P. Wildes\\	
        Lassonde School of Engineering\\
        Department of Electrical Engineering and Computer Science\\
        York University\\
}

\printtitle					
  	\vfill
\printauthor				
\hsp

\begin{nabstract}
\indent 
Video predictive understanding encompasses a wide range of efforts that are concerned with the anticipation of the unobserved future from the current as well as historical video observations. Action prediction is a major sub-area of video predictive understanding and is the focus of this review. This sub-area has two major subdivisions: early action recognition and future action prediction. Early action recognition is concerned with recognizing an ongoing action as soon as possible. Future action prediction is concerned with the anticipation of actions that follow those previously observed. In either case,
the \textbf{\textit{causal}} relationship between the past, current and potential future information is the main focus.
Various mathematical tools such as Markov Chains, Gaussian Processes, Auto-Regressive modeling and Bayesian recursive filtering are widely adopted jointly with computer vision techniques for these two tasks. However, these approaches face challenges such as the curse of dimensionality, poor generalization and constraints from domain specific knowledge. Recently, structures that rely on deep convolutional neural networks and recurrent neural networks have been extensively proposed for improving performance of existing vision tasks, in general, and action prediction tasks, in particular. 
However, they have their own shortcomings, \eg reliance on massive training data and lack of strong theoretical underpinnings.
In this survey, we start by introducing the major sub-areas of the broad area of video predictive understanding, which recently have received intensive attention and proven to have practical value. 
Next, a thorough review of various early action recognition and future action prediction algorithms are provided with suitably organized divisions.
Finally, we conclude our discussion with future research directions.
\newpage
\end{nabstract}

\setonehalfspace
\addcontentsline{toc}{section}{Table of Contents}
\tableofcontents
\listoffigures
\listoftables

\newpage
\startarabicpagination

\hsp

\chapter{Introduction}\label{CH1}

\section{Motivation}
In the computer vision community, video predictive understanding refers to the inference of the unobserved future from the current as well as historical information. 
From a theoretical perspective, video predictive understanding is of importance as it addresses the basic scientific question of how to extrapolate future events from previous observations. From a practical perspective, the ability to predict future events from videos has potential to provide positive impact for wide range of artificial intelligence applications, including autonomous vehicle guidance, effective time response to surveillance data and improved human computer interaction.

In video understanding, researchers are generally pursuing answers for two fundamental focal points: \textbf{what to predict} and \textbf{how to predict}. The intuition behind the first focus is that the future can be of several forms (\ie future semantic, physical trajectories, poses or pixel distributions) and it is non-trivial to explore its most feasible format to better assist real-life applications. To this end, diverse sub-areas have appeared that work on predicting the future in various knowledge domains. The second focus explores various mathematical techniques, information sources and data modalities to enhance the prediction accuracy.

\section{Overview of video predictive understanding}
The major sub-areas of video predictive understanding are pixel prediction, semantic segmentation prediction,  trajectory prediction, human motion prediction, predictive coding as well as early action prediction and future action prediction. While the focus of this review is early action prediction and future action prediction, in the next several paragraphs, each of these sub-areas will be briefly presented to provide context.

\textbf{Video pixel prediction} is concerned with generating pixelwise values for potential future images extrapolated temporally from previous observations. Early approaches to this problem attempted generation through learning procedures without adequate attention to basic elements of image structure (e.g., spatial gradients and optical flow) and yielded results that had undesirable artifacts (e.g., blurry images) as well as short prediction horizons \cite{ranzato2014video, srivastava2015unsupervised, oh2015action}. Subsequent work that explicitly considered such basic image structure yielded improved results \cite{mathieu2015deep, liu2017video}; however, prediction horizons were little extended. To support longer term predictions, work has relied on higher-level image analysis (e.g., body keypoint and motion analysis \cite{villegas2017decomposing, oliu2018folded, villegas2017learning}). Other work has made use of stochastic generative models to capture uncertainty \cite{lee2018stochastic, babaeizadeh2017stochastic, villegas2019high}. Yet other work has decomposed motion and appearance for separate processing via latent variables to better control outcomes \cite{tulyakov2018mocogan, hsieh2018learning, rybkin2018learning, jaegle2018predicting} as well as emphasized object centric motions \cite{ye2020object, wu2020future, ye2019compositional}. Generation of naturalistic future images beyond short time horizons remains challenging. Also, it is worth noting that such predictions may lend little insight into higher-level (e.g., semantic) understanding of future events.

\begin{figure}[htp]
    \centering
    \resizebox{.65\linewidth}{!}{\includegraphics[width=4cm]{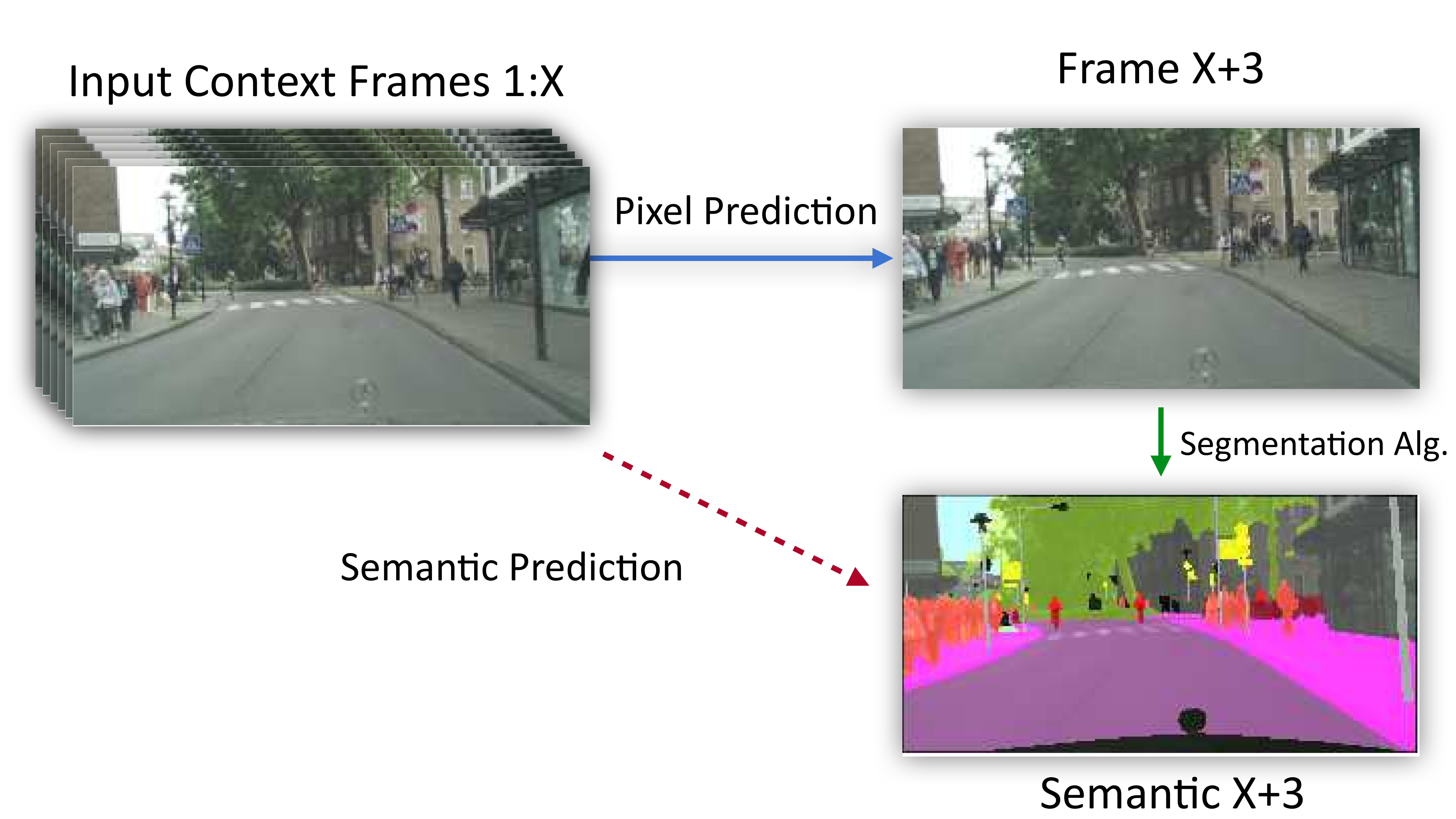}}
    \caption[Video pixel prediction in upper row and semantic segmentation prediction in lower row.]{Video pixel prediction in upper row and semantic segmentation prediction in lower row. A desired future scene parsing (semantic segmentation of a future frame) can be obtained via either: \textbf{1.} pixel prediction and then semantic segmentation; \textbf{2.} direct semantic prediction.}
    \label{fig:pixel_semantic}
\end{figure}

\textbf{Video segmentation prediction} bears high similarity with pixel prediction, yet it differs in that its outputs come in terms of pixelwise semantic labels (\eg road, tree, etc.). 
To achieve future semantic segmentations, one might either perform pixel prediction followed by standard semantic segmentation operations or directly infer future semantic segmentations. The latter approach is the focus of video semantic segmentation prediction. Figure \ref{fig:pixel_semantic} provides a visual illustration of the differences between pixel and semantic segmentation prediction. While different in output (i.e., single valued pixel labels vs. RGB values), semantic segmentation work has largely leveraged similar techniques as used for pixel predictions, with various levels of emphasis on purely appearance vs. dynamics semantics \cite{luc2017predicting, luc2018predicting, jin2017predicting, singh2018predicting, yagi2018future}. An important distinction is that the objective functions for semantic segmentation approaches typically rely on the downstream task (e.g., instance segmentation) compared to pixel predictions. 

\textbf{Video trajectory prediction} is another intensively investigated sub-area of video predictive understanding.
In this case, observed video is preprocessed into sequences of 2D coordinates that capture the trajectories of objects of interest (e.g., humans, vehicles). The goal is to predict a future sequence in the same format.
Early work focused on the forecasting of a single trajectory \cite{kitani2012activity}, whereas more recent research has expanded to the prediction of crowd trajectories \cite{alahi2016social, gupta2018social, sadeghian2019sophie, liang2019peeking, sadeghian2019sophie, ivanovic2019trajectron, salzmann2020trajectron++}. Among these efforts, some implicitly fused crowd information, by passing messages between targets, to better infer every single future trajectory \cite{alahi2016social, gupta2018social}. Other work actively embraced rich scene context to assist their predictions \cite{liang2019peeking, sadeghian2019sophie}. Yet other work explicitly used physical social distance to encode the crowd effect for their trajectory forecasting \cite{ivanovic2019trajectron, salzmann2020trajectron++}. Recently, a spatio-temporal version of a graph neural network has been found useful for this task \cite{mohamed2020social}. As with other areas of video prediction, long term prediction remains an outstanding challenge and while some work has made use of scene context, it is likely that much more can be explored in that direction. 

\textbf{Human motion prediction} focuses on foreseeing future human motion in the form of body key-points (\eg 24 body part coordinates \cite{martinez2017human}). This task is very akin to trajectory prediction, since its data is abstract key-points rather than raw pixels. The majority of work in this sub-area used recurrent neural networks in a Encoder-Decoder fashion to produce future pose sequences \cite{martinez2017human, fragkiadaki2015recurrent}. 
On top of that, a wide range of machine learning techniques have been applied to this sub-area, including attention \cite{tang2018long}, contrastive learning \cite{coskun2018human} and imitation learning \cite{wang2019imitation}. An important (and remaining) challenge in this area is to restrict generated motions to be realistic, even while varied. Advances along this direction have exploited geometric constraints \cite{gui2018adversarial} as well as generative modeling \cite{aliakbarian2020stochastic, wang2019imitation}.

\textbf{Video predictive coding} is rather different from the above sub-areas. It is concerned with coding raw video data into a high dimensional representation in such a way that it can be easily decoded into the future of input video. The learned representations are used as video features for various tasks. For example, early work along this line conducted video prediction learning with an objective to extrapolate frame-wise deep features that ultimately were applied to unsupervised action recognition \cite{srivastava2015unsupervised}. Follow on work enriched this idea with contrastive learning \cite{han2019video}. A recent effort that tried a similar idea on unsupervised video skeleton action recognition also observed success \cite{su2020predict}. Another effort found predictive coding meaningful for video retrievals \cite{yu2020encode}. 
A particularly interesting aspect of this line of work is that it requires no annotation, as the learning and feature distillation simply makes use of already available video data.
Researchers find this valuable since predictive coding is self-contained and requires zero annotations, thus it alleviates the heavy burden of human labeling.

\textbf{Action prediction} is principally concerned with the prediction of human actions. Among many important research topics in video understanding, action recognition arguably has received the most interest and efforts in the last decade \cite{kang2016review, kong2018human}. With tremendous advances achieved and various experiments explored in recent years, some researchers have refocused on the ability to \textit{foresee} the happening of certain actions rather than recognize the completed ones. The reason can be easily seen from the practical application point of view: Full action executions often take elongated time spans and thus performing the recognition algorithm after the action is completed would naturally lose the capability to react in real-time, which, however, in many conditions is essential.

\begin{figure}[h]
    \centering
    \resizebox{1.\linewidth}{!}{\includegraphics{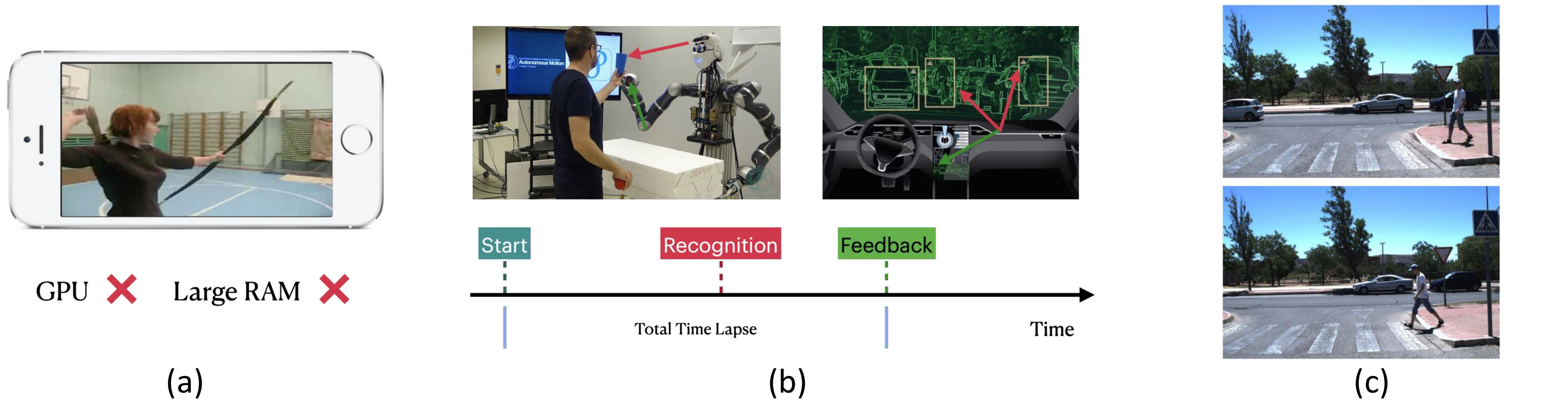}}
    \caption[Representative applications of early action recognition and future action prediction.]{Representative applications of early action recognition and future action prediction. Sub-figure (a)-(b) depict the use of early action recognition on reducing the computational budget of portable devices (\eg mobile phones typically lack GPU and memory space to process elongated videos) and on diminishing the response gap for fast feedback systems (\eg human-robot interactions and self-driving vehicles). Sub-figure (c) plots the application for future action prediction: Anticipating pedestrian behaviors at traffic intersections.}
     \label{fig:examples}
\end{figure}

As examples: In the video anomaly detection task, people expect a vision system that automatically spots any potentially disruptive events as early as possible and then sends out an alarm before any damage is done. Obviously, such a system is required to either recognize harmful actions in effective time or anticipate their happening in advance in space and time. Similarly, in the automonous driving setting, desirable unmanned vehicles need to be highly sensitive to pedestrian walking intentions, in particular at traffic crossings, to avoid accidents. To achieve that goal, the vehicle system needs to leverage contextual information from the scene (\eg traffic lights and crossings) together with pedestrian ego-motion, standing pose, heading angle, etc., to judge their most likely next action and respond in kind.

In general, based on application scenarios, researchers mainly explore two finer divisions: early action recognition and future action prediction. The former refers to recognizing the on-going actions as early as possible, while the latter to making plausible predictions on future actions that would happen after the execution of the current one. The focus of this review is on early action recognition and future action prediction. Figure \ref{fig:examples} provides illustrative examples of both early action recognition and future action prediction.

\section{Outline}
This chapter has served to motivate and provide a broad overview of video predictive understanding. Chapter \ref{CH2} will provide a detailed review of early action recognition. Chapter \ref{CH3} will provide a detailed review of future action prediction. Finally, Chapter \ref{CH4} will provide conclusions, including suggestions for future research directions.

\chapter{Early Action Recognition}\label{CH2}
\section{Overview}\label{Sec:early_actio}
\begin{figure}[htb]
    \centering
    \resizebox{.75\linewidth}{!}{\includegraphics{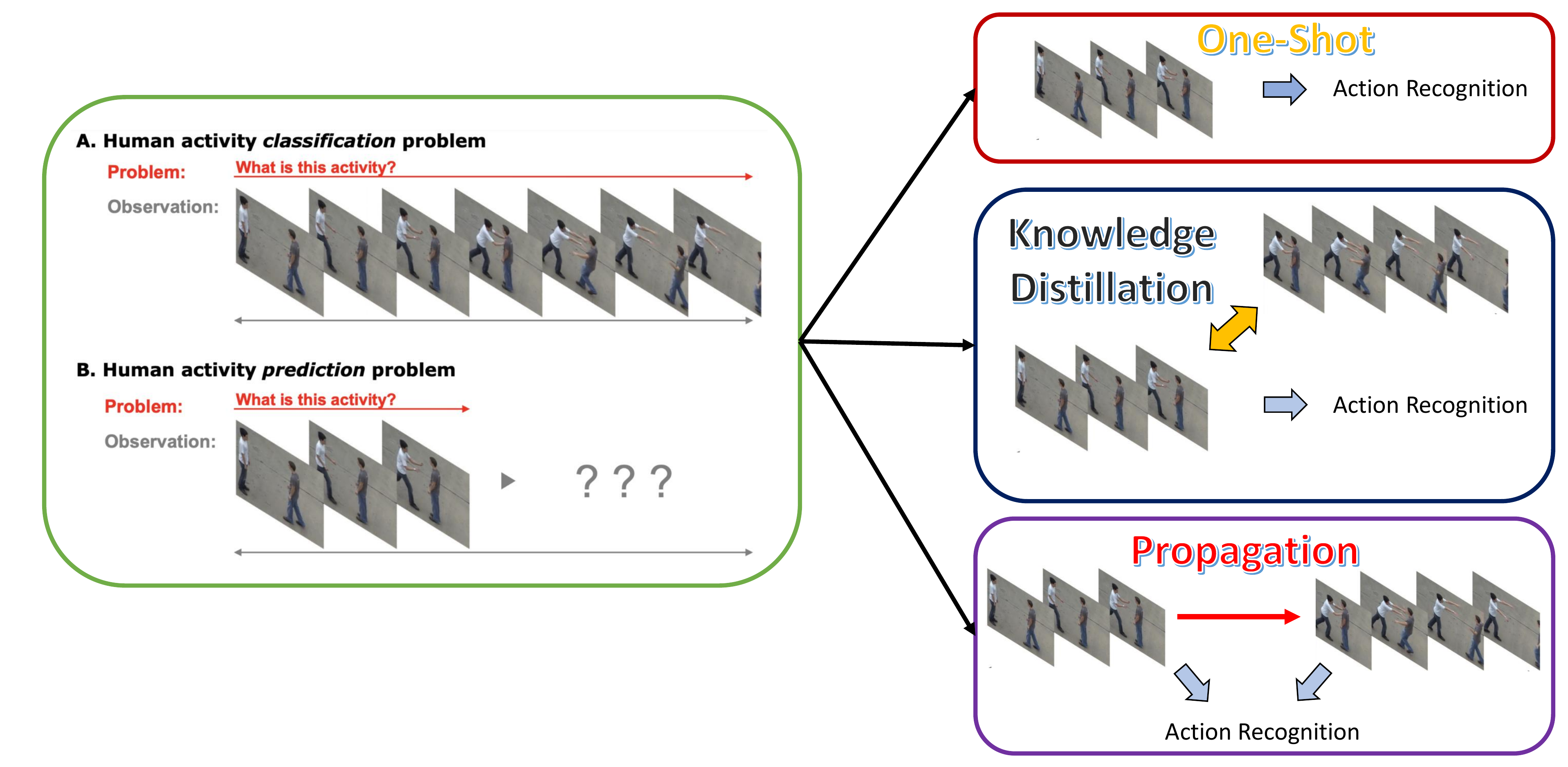}}
    \caption[The definition of early action recognition.]{(Left) The definition of early action recognition from \cite{ryoo2011human}. Preliminary observations of a particular ongoing action clip are used as input to generate the corresponding class label. Figure modified with permission from \cite{ryoo2011human}. (Right) Illustration of three major perspectives in current literature to tackle early action recognition: One-shot mapping, knowledge distillation and propagation.}
    \label{fig:action_prediction}
\end{figure}

One extensively investigated sub-area of video predictive understanding is early action recognition. Its core mission is to recognizing actions while they are still evolving and predicting the correct action labels as early as possible, as illustrated in Figure \ref{fig:action_prediction}.

Formally, the problem can be defined as: Given a sequence of action video frame observations $y_{1:T}$ and its semantic label $\mathcal{X}$, where $1$ and $T$ represent the starting and ending indices of action frames, define a function $f : y_{1:t} \rightarrow \mathcal{X}$, which maps various partial segments $y_{1:t}$ to the groundtruth label domain. In particular, it is favorable to have high recognition performance on relatively small frame index (\ie $t \rightarrow 0$), meaning that the algorithm is able to recognize actions correctly with very limited initial observations.

Apparently, there is a fundamental assumption behind this problem setting: The full action sequence supports superior recognition compared to observation of only an initial segment, otherwise the problem is unnecessary. To validate this assumption, the original advocate of this research direction, M. S. Ryoo, conducted a series of experiments that operated existing action recognition algorithms on various partial action clip segments (\ie $y_{1:t}, t \in (1, \ldots, T)$) and demonstrated that insufficient observations in action temporal scope resulted in much inferior recognition accuracy \cite{ryoo2011human}. Also, as a visual example, the partial observation in Figure \ref{fig:action_prediction} (bottom row) can be sensibly categorized into \textbf{pushing} or \textbf{hugging} with equal chance, whereas the full action sequence (top row) reveals high likelihood for \textbf{pushing}.

Follow on work defined a systematic evaluation procedure for early action recognition \cite{kong2014discriminative}: Uniformly divide a full video $y_{1:T}$ into $K$ segments $y[\frac{T}{K}(k-1) +1, \frac{T}{K}(k)]$ where $k = 1, ..., K$ is the index of the segment. The length of each segment is $\frac{T}{K}$ and different videos usually have separate segment lengths. Thus, a partial observed video sequence can be described as $y[1, \frac{T}{K}(k)]$ and recognition accuracy at all possible $k$ should be reported (\eg they pre-define $K=10$ and report 10 sets of accuracies). The majority of successive work follows this paradigm.

In the last decade, researchers approach this task from various perspectives and hereby we group them into three major divisions: \textbf{1.} formulating one-shot mappings from partial observations to groundtruth labels of full observations; \textbf{2.} distilling the information from the full action clip observations into partial observations; \textbf{3.} propagating the limited partial information into the future in a temporal extrapolation fashion. In the following sections, we review all recent research according to these divisions.

\section{One-Shot Mapping Based}
\begin{figure}[ht]
    \centering
    \begin{minipage}{.52\textwidth}
        \centering
        \includegraphics[width=1.\textwidth]{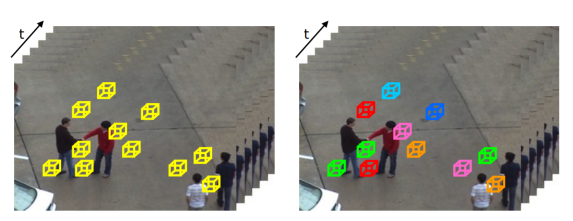}
        \caption*{(a)}
    \end{minipage}
    \begin{minipage}{.38\textwidth}
        \centering
        \includegraphics[width=1.\textwidth]{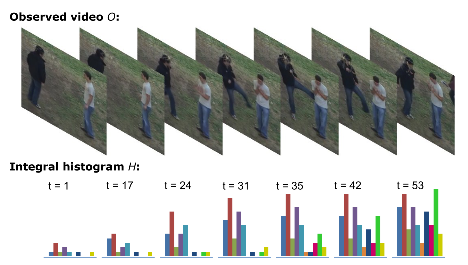}
        \caption*{(b)}
    \end{minipage}
\caption[STIP features from videos and clustered results by color for each frame]{(a) STIP features from videos and clustered results by color for each frame. (b) Demonstration of Integral Bag-of-Word (IBOW) used in \cite{ryoo2011human}. For each time step, its feature histogram amasses all previous observations. Figure reproduced with permission from \cite{ryoo2011human}.}
\label{fig:stip_bow_hist}
\end{figure}

The basic assumption underlying one-shot mapping based approaches is that a partial observation of an action video provides sufficient information in and of itself to define the appropriate overall action class regardless of the unobserved clip portions. What appears to be the first work along these lines \cite{ryoo2011human} made use of sequential feature matching with an Integral Bag-of-Word (IBoW) representation \cite{sivic2008efficient}. The intuition being that the feature distributions of partial and complete action clips should be similar, if they derive from the same action. To avoid inefficiencies that would result from training classifiers for each temporal segment, they initially extract Spatio-Temporal Interest Point features (STIP \cite{laptev2005space}) from raw videos and re-organize them into visual word histograms (IBoW) according to color distributions. Then, a Gaussian kernel matching function is used to generate a probability score representing the likelihood between partial and whole observations. Finally, the selected class label is that of the full action that produces the largest likelihood. As illustrated in Figure \ref{fig:stip_bow_hist}, for each action segment, a color histogram is calculated. Similar work improved the feature representation as well as selection with sparse coding but shared the overall structure \cite{cao2013recognize}. 

While the IBoW approach supports fast inference and is learning free, it operates under two restrictive assumptions. First, the feature extractor, STIP, yields an appearance based feature descriptor (indeed STIP comes from spacetime Harris-corner detection \cite{harris1988combined}), which is sensitive to view changes, illumination and other nuisance details of image acquisition. Thus, targets from the same action category are assumed to be similar in appearance. Second, the scheme to matching feature histograms between partial and full actions is not always practical since actions can evolve dramatically, to the extent that the beginning appears totally unlike the ending.

Subsequent work following essentially the same framework improved on the original by employing stronger features as well as more sophisticated machine learning-based classifiers \cite{lan2014hierarchical}. More specifically, the feature representation was enhanced by merging motion descriptors (iDT \cite{wang2013action}) with appearance descriptors (HOG \cite{dalal2005histograms}). Furthermore, to capture the human movements at all levels, they proposed an approach to capture features at three hierarchical temporal scales: a coarse level feature that processes generic videos by taking frames as a whole; a mid-level feature that captures viewpoint specific information by extracting features only belonging to certain specific actors (\eg \textit{level2}, focuses on regions that are pre-clustered by different human actor appearances); a fine-grained movement level that captures pose specific information (\eg \textit{level3} extracts features only from regions that share the same actor appearance \textit{and} pose type, which is again obtained by clustering). To finally assign class labels to given video clips, the authors adopted Support Vector Machines (SVMs) \cite{cortes1995support} that were trained on features from all three levels to jointly make the decision.

Though this relatively new and detailed approach advanced its predecessors through adopting stronger features, hierarchical extractions and learning-based classifiers, it still relied on the assumption that fine grained analysis of an initial partial observation can bridge the past and future. A notable downside to this particular extension is that it relied on clustering algorithms for grouping actors and poses \cite{lan2014hierarchical}, which can be noisy (\eg it is sensitive to the color of clothing).

Subsequent work \cite{singh2017online} disposed of unsupervised clustering for detection of actors and their poses by instead making use of object/human detectors pretrained on massive datasets (\eg SSD \cite{liu2016ssd}). Therefore, they proposed to jointly perform action localization as well as action early recognition, based on the assumption that actions only associate with behaviors of human actors, and called the detected actor regions \textbf{action tubes} (\ie bounding box coordinates of detected actors). For given frames, this approach generates bounding box regions together with class labels for all possible regions within each frame 
and rejects negative bounding boxes by matching its Intersection Over Union (IOU) with selected regions from previous frames. Since the approach can consume frames online, early action recognition is naturally decided by the classification result from contemporary observations. 

Arguably, human active regions (\eg \cite{lan2014hierarchical, singh2017online}) are still very coarse if being compared with human skeleton analysis \cite{xia2012view} where each body part (\eg arms, elbows, shoulders etc.) is spatially spotted and used. Correspondingly, other work \cite{chen2018part} has suggested to first use human body key-point extraction tools to locate an actor's body structure information and select the salient body feature parts for early action recognition, with a reinforcement policy learning \cite{mnih2015human} function for performing the selection. 
Along similar lines, research using 3D body skeleton analysis for early action recognition has focused on adaptively selecting the best historical information window for predictions in data-driven fashion \cite{liu2018ssnet}.

\begin{figure}[htb]
    \centering
    \resizebox{.8\linewidth}{!}{\includegraphics[width=4cm]{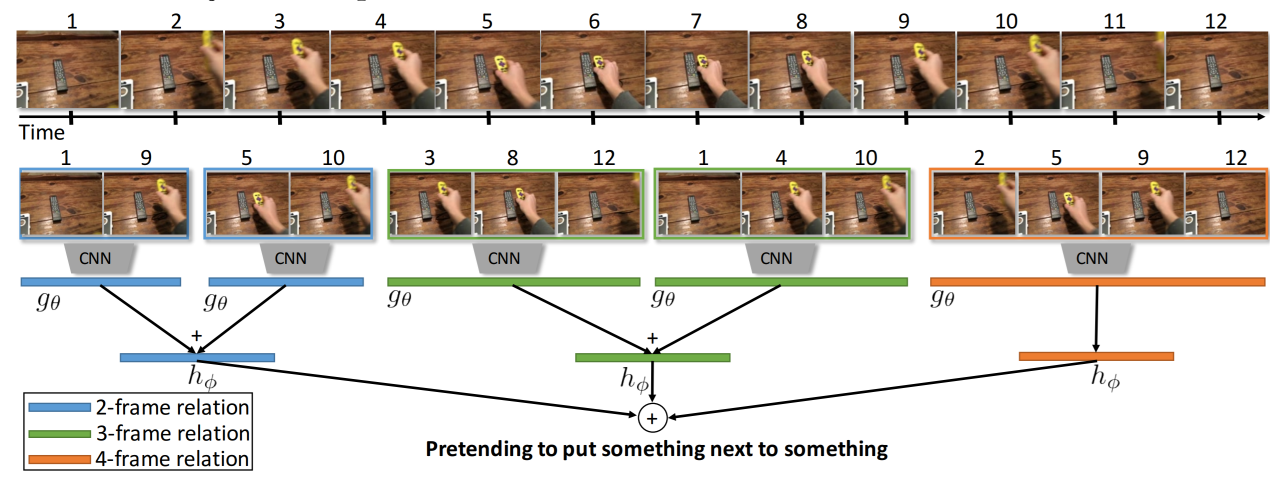}}
    \caption[Example of temporal relation reasoning for early action recognition.]{Example of temporal relation reasoning through sampling frames with multi-scale time intervals \cite{zhou2018temporal}. Frame group $(1,9)$ denotes the relation \textbf{put next to} whereas group $(1, 4, 10)$ denotes the \textbf{take-away} relation. Considering all sub-relations leads to the inference of the action label. Figure reproduced with permission from \cite{zhou2018temporal}.}
    \label{fig:temporal_reasoning}
\end{figure}

Notably, the above reviewed works are mostly actor centric, i.e., they base their inference on analysis of the actor alone, with little regard for other objects and scene context. In contrast, a recent effort has instead considered object centric human actions (e.g., fetching objects or moving something to somewhere) \cite{zhou2018temporal}. This work reckoned that the key to reasoning about complex activities is the pattern of temporal developments between subactions and/or objects at various scales, something that was not modeled explicitly in previous work.
In particular, the authors built a ConvNet to perform feature extractions and 2-layer MLPs to capture frame-wise temporal relations. More specifically, multiple sets of frames sampled from various temporal scales are combined together as the final representation to classify. For example, the collection of frame index groups $(1,9)$, $(2, 5, 9, 12)$ and $(1,4,10)$ represent three sets of temporal relations where states of object or actor (\ie human hand in Figure \ref{fig:temporal_reasoning}) transit differently in both speeds and conditions. 

While the above approach reasons about temporal relations, it aggregates frame-wise information globally and thereby does not exploit spatial relations of elements within a single frame. More recent work has extended this principle to the spatial dimension \cite{sun2019relational}. Similar to \cite{singh2017online}, it used a learning based detector to generate actor proposals for all characters in one frame. The relation between multiple actors is modeled as nodes and edges as in a graph \cite{santoro2017simple} and a Graph Neural Network (GNN) \cite{hamilton2017representation} is adopted to diffuse information among nodes. Following this direction, a multi-scale spatio-temporal graph based method further explores fine-grained human-to-human and human-to-object interactions for this task \cite{wu2021spatial}.

Finally, some other work considers early action prediction as a special case of action recognition in that only a few action clips that are easily differentiable under full observations become difficult when incomplete. To deal with such difficult cases, they equipped the regular recognition framework with an extra memory module to log hard examples explicitly \cite{kong2018action}.

\section{Knowledge distillation Based}
Methods in the previous section mainly focus on reasoning about the future directly from the past. The future unobserved information provides negligible contribution to the partially observed information, except the overall action label being used for training. Here, it is worth remembering that previous work has shown that full observation almost always excels over the partial and no one has forbidden the use of unobserved information in training \cite{ryoo2011human}. The more urgent question concerns \textit{how} to use such unobserved information.
Hereby, we discuss research that attempts to lend power from unobserved data in training in order to either enrich the feature representation of partial data or encourage the classifiers to recognize partial data with ease.
\begin{figure}[htb]
    \centering
    \resizebox{.7\linewidth}{!}{\includegraphics[width=4cm]{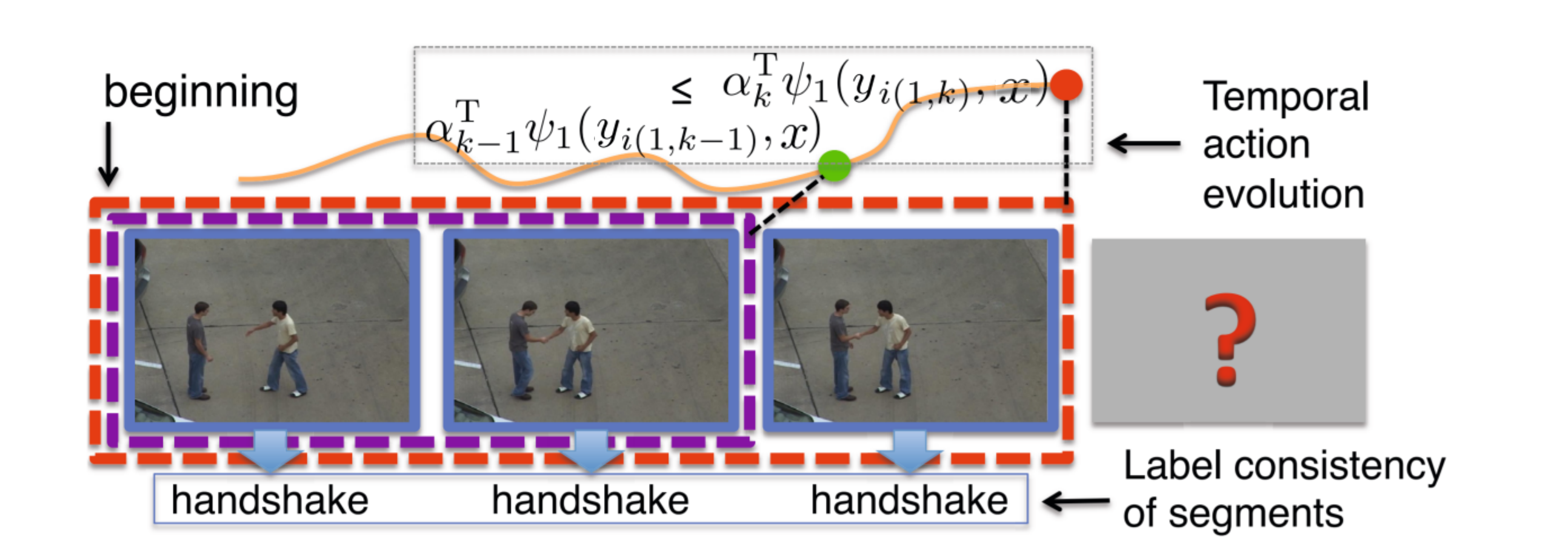}}
    \caption[Temporal action evolution patterns for early action recognition.]{
   Temporal action evolution patterns can be enforced through a score function \cite{kong2014discriminative}, as given in Equation \ref{eq:kong14}. Classifiers may work better with partial videos by considering its ranking score with full videos. Figure modified with permission from \cite{kong2014discriminative}.
   }
    \label{fig:kong2014}
\end{figure}

Initial work that informally considered mutual information in early action recognition trained video segment classifiers that align the temporal order of partial observations to training data \cite{kong2014discriminative,kong2015max}. They achieved this result by using a Structured SVM (SSVM) \cite{tsochantaridis2005large}. More specifically, they evenly divided action videos into equal-length segments, extracted features from each segment, integrated segments into one chunk, such as $y_{1:k}$, and trained SSVM classifiers according to the following constraints (without loss of generality, two progress levels ($k-1, k$) is given):
\begin{equation}
    \alpha^{T}_{k-1}\psi_{1}(y_{1:k-1}, x) \leqslant \alpha^{T}_{k}\psi_{1}(y_{1:k}, x),
\label{eq:kong14}
\end{equation}
where $y_{1:k}$ is the video segment feature accumulated from the very begining till the $k_{th}$ level, $x$ is the groundtruth label, $\alpha^{T}_{k}$ is the score matrix for progress level $k$ and $\psi_{1}(y_{1:k}, x)$ is a joint feature map that represents the spatio-temporal features of action label, $x$, given a partial video $y_{1:k}$. The authors assumed 10 progress levels, such that $k \in (1, 10)$.

The above formulation emphasized temporal label consistency: Besides assigning a correct class label to given input videos, classifiers also need to capture the temporal evolution of actions by monotonically increasing the score for segments across time (first seen in early event detection \cite{hoai2014max}). The intuition coincides with daily human experiences: As we see more, we should know better than before. Some other work follows a similar intuition and built their early activity detection model based on ranking loss \cite{ma2016learning}. It also has been proposed to incorporate global human intentions as a complimentary learning objective to assist partial observation learning, but for action prediction in first-person-view videos \cite{ryoo2015robot}. One particular work further argued for assigning various weights for features from every progress level and suggested using temporal saliency to locate the corresponding weights \cite{lai2017global}. Also, other work adopted the same monotonicity constraints and grounded the feature aggregation on human poses \cite{raptis2013poselet}. Finally, an effort focused on refining the classification confidence score at each time-step in an incremental manner \cite{hou2020confidence}. That work relied on an attention mechanism to provoke high confidence on early observations.
Notably, all these efforts differ from methods in the previous section in that unobserved data gets involved to empower the partial data during training.

\begin{figure}[htb]
    \centering
    \resizebox{.65\linewidth}{!}{\includegraphics[width=4cm]{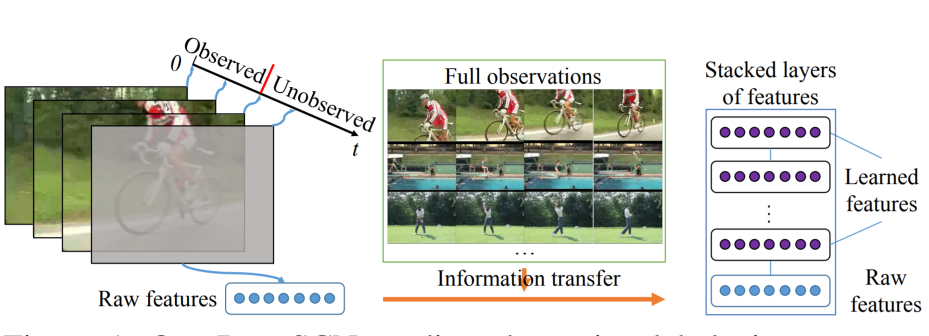}}
    \caption[Illustration of the feature transformation approach.]{Illustration of the feature transformation approach of \cite{kong2017deep}. Raw incomplete features are transformed into features from the fully-observed to gain more discriminative information. Figure reproduced  with permission from \cite{kong2017deep}.}
    \label{fig:deep_scn}
\end{figure}

Yet another approach to using a scoring function was proposed by the originators of this direction for early action recognition \cite{kong2014discriminative}. In this alternative approach the attempt was made to infer complete future features from partial observations \cite{kong2017deep}, something not previously attempted in early action recognition. Their work supposes that we have features of partial observation, $g(y_{1:t})$, and full observation, $g(y_{1:T})$ ($t \in (1, \ldots, T)$), in the training stage, and proposed to reconstruct the full from partial according to:
\begin{equation}
    \operatorname*{argmin}_{W, \theta} || g(y_{1:T}, \theta)  - W g(y_{1:t}, \theta)||^{2},
\label{eq:deepscn}
\end{equation}
where $W$ is the feature transformation matrix and $g(y, \theta)$ is the feature extraction function with parameter $\theta$. As shown in Figure \ref{fig:deep_scn}, raw partial observations (\eg riding on trails with green grass background) could be mixed with any other actions that share the similar background context (\eg golfing or swimming in pools aside the lawn). Transforming the partial into full observations at the feature level helps differentiate between the possibilities (\ie capturing more details of the bicycle structure as well as human riding motions).

In implementation, the state-of-the-art 3D CNN video features (\ie C3D \cite{tran2015learning}) were employed for $g$ and the transformation matrix, $W$, was learned. A downside of this approach is that multiple transformation matrices needed to be learned, i.e., one for each $k$. A more recent upgrade on this work \cite{kong2018adversarial} overcame this concern by converting the deterministic feature transformation process \cite{kong2017deep} to a conditional generative process, with an adversarial learning component \cite{goodfellow2014generative} to promote realism. The progress level, $k$, then plays as a hyper-parameter to inform the regressor (emboddied as a ConvNet) about the input type and correspondingly adjust its weights for the desired output. 
A subsequent development argued for alleviating the heavy burden of high-dimensional feature regression \cite{qin2017binary}. The authors suggested transforming features into a low dimension binary code, as seen in Hashing encoding theory \cite{calonder2010brief}, and then operating the feature reconstruction on the binary encoded counterpart.

\begin{figure}[htb]
\centering
\begin{minipage}{.55\textwidth}
    \centering
    \includegraphics[width=.8\textwidth]{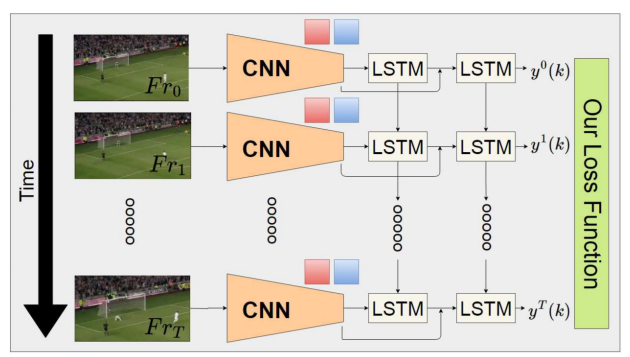}
\end{minipage}
\caption[Algorithmic diagram of Encourging-LSTM.]{Algorithmic diagram of \cite{sadegh2017encouraging}. Sequential frames are processed by deep neural networks step by step and a time-variant cross-entropy loss function encourages early recognition. Figure reproduced with permission from \cite{sadegh2017encouraging}.}
\label{fig:encourage_lstm}
\end{figure}

Other work approached the same problem by encouraging classifiers to assign correct class labels as early as possible \cite{sadegh2017encouraging}. They noticed that early work also made efforts on regularizing the classifiers \cite{kong2014discriminative}; however, none had explicitly enforced the early recognition of actions. In their framework, a sequence of video frames $(y_0, ..., y_{T-1}, y_{T})$ (denoted as $Fr_{t}$ in Figure \ref{fig:encourage_lstm}) is processed by a ConvNet extractor, LSTM sequential aggregation and MLPs as the final classifier, which would regress a sequence of class probability for certain class index $c$ as: $(\hat{x}_{0}(c), \hat{x}_{1}(c), \ldots, \hat{x}_{T}(c))$ (denoted as $y^{t}(k)$ in Figure \ref{fig:encourage_lstm}).

They designed a novel cross-entropy loss to scale the false-positive loss value based on the incoming time index according to
\begin{equation}
\begin{aligned}
  \mathcal{L}(x, \hat{x}) &= -\frac{1}{C} \sum^{C}_{c=1}\sum^{T}_{t=1} [x^{t}(c)log(\hat{x}^{t}(c) + \\
  &\frac{t(1-x^{t}(c))}{T}log(1-\hat{x}^{t}(c))],
\end{aligned}
\label{eq:time_ce_loss}
\end{equation}
where $C$ denotes the total number of action classes.

Recalling that the false negative term (\eg first term in Equation \ref{eq:time_ce_loss}) aims for correctly assigning labels, whereas the false positive term (second term) for suppressing wrong labels, the authors made the relative weights for the former higher than the latter at the beginning of the sequence. That means high probability score for incorrect class labels is allowed as long as the score for correct ones is high enough. The key intuition behind this choice is tolerance for ambiguity at early stages. Therefore, the overall loss encourages the correct recognition in early time stamps.

Parallel research also exploited the property of loss functions for early action recognition \cite{hu2018early}. Instead of generating hard labels (\eg SVMs only output one fixed label and discard other possibilities), they proposed to produce soft labels in a regression fashion, rather than classification. As a concrete example, they map input videos $y_{1:t}$ using transformation matrix $W$ to a vector $x$ that contains scores for all possible classes $C$ as:
\begin{equation}
    x = s(t) W g(y_{1:t}, \theta), x \in \mathcal{R}^{C},
\label{eq:soft_label_regression}
\end{equation}
where $g(y_{t}, \theta)$ is the feature extraction function, output $x$ has dimension $C$ and $s(t)$ is a scaling factor that monotonically increases along the time index. 
The authors adopted a real value time index, $t$, rather than the evenly allocated progress level, $k$, to enable online inference.
Arguably, extensively adopted softmax classifiers have enabled probability scores for all classes and other related work has revealed its effectiveness \cite{sadegh2017encouraging}. However, no direct comparison of the performance between soft-regression and softmax classification was provided.
\begin{figure}[htb]
    \centering
    \resizebox{.45\linewidth}{!}{\includegraphics{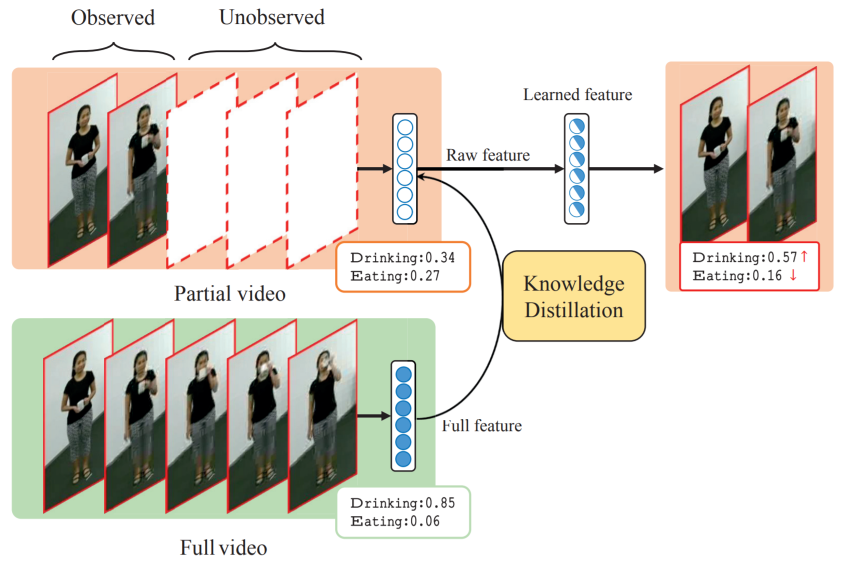}}
    \caption[Illustration of the teacher-student learning approach.]{Illustration of the teacher-student learning approach. The student branch takes partial action clips as input (\eg top row), while the teacher branch (bottom row) takes the full clip. Knowledge distillation implemented through feature distance minimization encourages the student branch to learn from the teacher. Figure reproduced with permission from \cite{wang2019progressive}.}
    \label{fig:teacher_student}
\end{figure} 

Very recently, knowledge distillation \cite{hinton2015distilling} has been applied to squeeze fully-observed information into partial observations \cite{wang2019progressive}.
Two neural networks were employed: One acts as a teacher that gets access to full observations; the second acts as a student with only partial access. The student's representation is driven toward that of the teacher  by minimizing the Mean Square Error (MSE) between the latent features and the Maximum Mean Discrepancy (MMD) between the predicted classification probabilities; see Figure \ref{fig:teacher_student}. Some follow on efforts emphasized both the feature reconstruction consistency and semantic knowledge consistency via integrating multiple loss functions (\eg feature distance loss, adversarial loss and classification loss) for a joint optimization \cite{xu2019prediction, wang2019early}. Indeed, it seems that the distance metric plays a critical role in building the connection between video parts. Further, one recent work has found significant improvements via adopting the \textbf{Jaccard} similarity metric to learn the transition between partial and full observations \cite{fernando2021anticipating}.

\section{Propagation Based}
Another way to exploit future information is by propagating the partial observation into the future. Typically, this type of approach extrapolates the visual representations into the future and then applies an existing classifer. 
The key difference between propagation and mutual information is whether or not the system performs \textbf{temporal extrapolations}. Mutual information methods from the previous section emphasize how to enrich partial observations with future information, whereas propagation based methods in this section focus on how to generate an accurate future based on current observations. Mathematically, propagation based approaches start from any incomplete input $y_{1:t}$, assuming $t < T$, and produce the subsequent information $(y_{t+1}, y_{t+2}, ..., y_{T})$; thus, they can be considered as temporal extrapolation. In contrast, during either the testing or training phase, mutual information based approaches do not extrapolate.
Classification is then based on the the concatenation of the observed and extrapolated.

An early exemplar of the propagation approach operated by working with relatively high-level visual representations so that they can be readily processed by extant recognition tools \cite{vondrick2016anticipating}.
\begin{figure}[htb]
    \centering
    \resizebox{.6\linewidth}{!}{\includegraphics{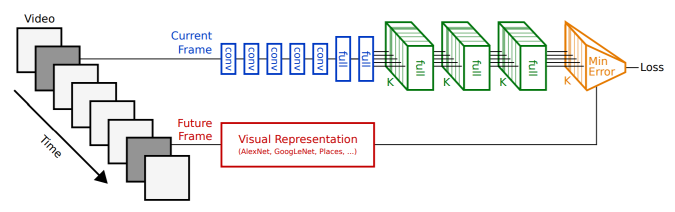}}
    \caption[Deep regression network from with mixtures of output layers for action prediction]{Deep regression network from \cite{vondrick2016anticipating} with mixtures of output layers. Blue boxes are convolutional and fully connected layers for unified processing on each frame, while green boxes represent the mixture of output layers. To push the predicted visual representation close to the real future, MSE loss (\ie yellow box) is calculated with the groundtruth extracted from future frames in a self-supervised way. Figure reproduced with permission from \cite{vondrick2016anticipating}.}
    \label{fig:anticipation}
\end{figure}
In practice, they formulated a deep regression network, consisting of multiple convolutional and fully connected layers, to temporally regress the future high-level deep feature (\ie fc7 activation output from AlexNet \cite{krizhevsky2012imagenet}). To generate the label for future representations without supervision, they collected a small group of labeled video representations, calculated the similarity score and chose the best one. In order to model multi-modality, they extended the original single output structure to $K$ mixtures (\eg denoted as green boxes in Figure \ref{fig:anticipation}). 

This approach has two positive attributes. First, to the extent that the deep features capture semantically meaningful representations, they should be well suited to support recognition. Second, the mixture output helps to model the possibly multimodality of future outcomes. A notable downside is the potentially prohibitive dimensionality of the representation. Nevertheless, it unveiled a novel approach and inspired several follow-on efforts. A subsequent work used Recurrent Neural Networks (RNNs) with learnable Radial Basis Function (RBF) kernels to improve the nonlinearity of predictions \cite{shi2018action}. 
Yet other work upgraded the deep regression component with a Markov Decision Process (MDP) as well as an RNNs \cite{zeng2017visual}, but shared the same intuition.

\begin{figure}[ht]
    \centering
    \begin{minipage}{.48\textwidth}
    \includegraphics[width=1.\textwidth]{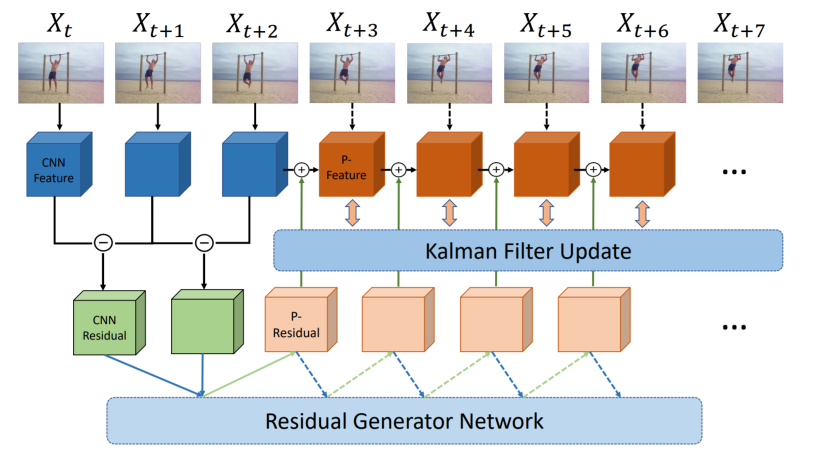}
    \end{minipage}
    \begin{minipage}{.48\textwidth}
    \includegraphics[width=1.\textwidth]{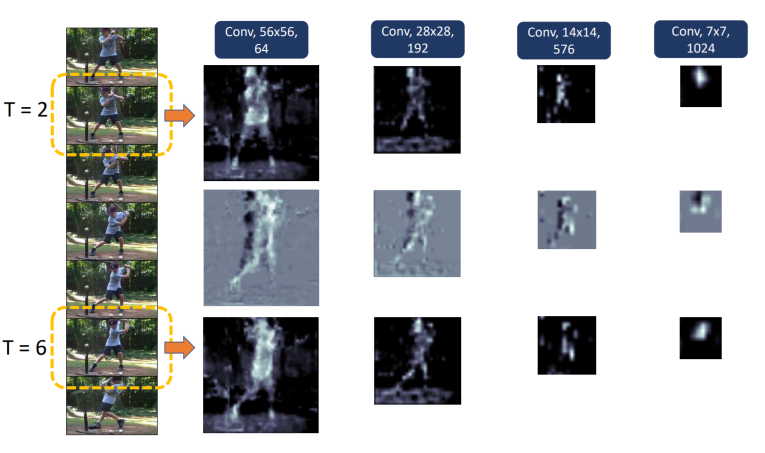}
    \end{minipage}
\caption[Deep Kalman filtering.]{ The left sub-figure depicts the overall workflow from \cite{zhao2019spatiotemporal} that revisited Kalman filtering under deep network structure: Partial observations (\ie blue boxes) and their pair-wise residuals (\ie green boxes) are used to propagate information into the future. Kalman filtering update is performed at each time step to combat error accumulation; The right sub-figure shows an ablation study that examines different feature stages through visualization. The intermediate feature stage (\eg $3^{rd}$ column from left) performs the best since it keeps critical information of actors while suppresses noisy background present
in the shallow layer. In comparison the vectorized features deeper in the network are too coarse and suppress important details. Numerical studies backed-up the visualizations. Figure reproduced with permission from \cite{zhao2019spatiotemporal}.}
\label{fig:iccv19_he}
\end{figure}

While the initial studies on propagation-based early action recognition achieved promising results, critical questions were left unanswered: What type of feature works best for propagation? How can video redundancy be dealt with? How can remote predictions avoid degeneracy? Further, design decisions appeared to be made heuristically without support from ablation studies. These concerns were addressed in an approach that made use of classic recursive filtering coupled with more careful design \cite{zhao2019spatiotemporal}; see Figure \ref{fig:iccv19_he}.

They proposed to propagate the frame-wise feature residuals to lessen the temporal redundancy and thereby forced the model to focus on the underlying dynamics, rather than static background. Moreover, to combat error accumulation, Kalman filtering is incorporated into a deep network, which iterated the \textbf{prediction} and \textbf{update} steps to stablize long-term predictions. Further, they provided an answer to the above question regarding proper features by examining various feature layers in a deep network in propagation and determined that the intermediate layer features, rather than vectorized last layer feature typically used in earlier work, is more suitable for propagation. Presumably, this intermediate level of representation provides the right level of abstraction from the raw video without loosing too much detail.

Other work combined the idea of feature propagation with a reinforcement reward function to encourage early recognition \cite{gao2017red}; Yet other work propagated both the RGB and motion (\ie optical flow) features from deep networks and adopted adversarial learning to boost the realism of their outputs \cite{gammulle2019predicting}. 
In contrast to most recent work that heavily used deep features, an effort has employed hand-crafted Dynamic Image (DI) \cite{bilen2017action} features for propagation, but still depends on deep recognition models for the label assignment \cite{rezazadegan2017encoding}. Finally, an approach that propagated both graph node features (\eg 2D coordinates) and structures (\eg adjacency matrices) has been proposed to tackle a recently appeared task, namely group-player early action recognition \cite{chen2020group}.

\section{Datasets and Performance}
In this section, we describe datasets, evaluation metrics and performance for the majority of early action recognition approaches discussed in this section. As all datasets are used for both action recognition and early recognition, comprehensive comparisons of them can be found elsewhere \cite{kong2018human, kang2016review}.

\textbf{Evaluation} on early action recognition is based on the same recognition accuracy metric as in action recognition, but with a focus on the \textbf{observation ratio} $K$ ranging from $10\%$ to $100\%$, as discussed in the beginning of this chapter. Since each video only owns a single label, the video level mean average precision is reported against each $K$ value. 

\textbf{UCF-101} \cite{soomro2012ucf101} is a dataset collected from a real-life video platform (Youtube) and trimmed for action recognition (each video contains exactly one action). It includes 101 distinct action classes and 13,320 overall video clips with at least 100 videos for each category. UCF-11 and UCF-50 are two earlier versions of the same dataset that contain 11 and 50 action categories respectively. One critical issue of UCF-101 is that videos from YouTube could be very biased by low-level features, meaning low-level features (\ie color and gist) are more discriminative than mid-level features (\ie motion and shape). Researchers have been progressively evaluating both action recognition and early action recognition on UCF-101.

\textbf{HMDB-51} \cite{kuehne2011hmdb} is a large scale human action recognition dataset that comprises 51 daily action categories. It is worth noticing that it contains some fine-grained human facial motions, such as smiling, laughing, chewing etc, in static background windows, which are not seen in other comparable datasets. This dataset also encompasses a wider range of camera motions and viewpoint changes compared to UCF-101. In total, there are a total of 6,766 video clips with at least 102 videos for each class. There are three official data splits and the averaged evaluation score is often reported as final results.

\textbf{JHMDB-21} \cite{jhuang2013towards} is a subset of HMDB-51 with a special focus on 21 human body joint action categories where additional annotations on joints (13 2D-coordinates body joints) are provided. Compared with HMDB-51, categories that mainly contain facial expressions (e.g. smiling), interaction with others (e.g. shaking hands), and very specific actions (e.g. cartwheels) were excluded. Therefore, there are a total of 928 video clips.

\textbf{BIT} \cite{kong2012learning} consists of 8 classes of human interactions (bow, boxing, handshake, high-five, hug, kick, pat, and push), with 50 videos per class. Videos are captured in realistic scenes with cluttered backgrounds, partially occluded body parts, moving objects, and variations in subject appearance, scale, illumination condition and viewpoint. Even though BIT has a relatively limited number of classes and videos, it is a complex dataset in that: \textbf{1.} The backgrounds as well as actor appearances are highly non-discriminative. Various categories often share the same background and actors; \textbf{2.} The starting and ending phases of the videos are highly similar (\ie actors standing still and facing each other). Thus, researchers often have to resort to motion based features for comparable performance.

\textbf{UT-Interaction} (UTI) \cite{ryoo2015ut} is comprised of 2 sets with different environments. Each set consists of 6 types of human interactions: \textit{handshake, hug, kick, point, punch and push}. Each type of interaction contains 10 videos, to provide 60 videos in total. Videos are captured at different scales and illumination conditions. Moreover, some irrelevant pedestrians are present in the videos.

\textbf{TV Human Interaction} \cite{patron2010high} consists of 300 video clips collected from over 20 different TV shows. It contains five action classes: \textit{handshake, high five, hug, kiss and none}. The class “none” represents all other more general actions such as walking and standing. Annotations are provided for every frame of the videos, including the upper body bounding boxes, discrete head orientations and action labels for each person. 

\textbf{Something2something} \cite{goyal2017something} is a dataset that shows human interaction with everyday objects. In the dataset, humans perform a pre-defined action with a daily object. It contains 108, 499 video clips across 174 classes. The dataset enables the learning of visual representations for physical properties of the objects and the world. Compared with others, the dataset shows primarily human related motions (mostly only hand motions are visible). 

\textbf{SYSU-3DHOI} \cite{hu2015jointly} (3D Human-Object Interaction) includes 480 RGBD sequences from 12 action categories, including “playing phone”, “calling phone”, “pouring”, “drinking”, etc. For building this set, 40 participants were asked to perform 12 different activities freely. For each activity, each participant manipulates one of the six different objects: phone, chair, bag, wallet, mop and besom. Therefore, there are in total 480 video clips collected in this set. The contained activity samples have different durations, ranging from 1.9s to 21s. For each video clip, the corresponding RGB frames, depth sequence and skeleton data were captured by a Kinect camera \cite{zhang2012microsoft}.

\textbf{NTU-RGBD} \cite{shahroudy2016ntu} is by far the largest public set for 3D action recognition and prediction. It contains more than 56,000 video samples with about 4 million frames from 60 action categories. All of these action samples were recorded by a Kinect camera \cite{zhang2012microsoft} from three different views. For collecting this set, 40 subjects were asked to perform certain actions several times. This set is very challenging for early action prediction mainly due to its larger scales of quantity, greater diversity in action categories and more complexity in human-human interaction and human-object interaction.

\textbf{ORGBD} \cite{yu2014discriminative} (Online RGB-D) was collected for online action recognition and early action prediction. There are seven types of actions that people often do in the living room: drinking, eating, using laptop, picking up phone, reading phone (sending SMS), reading book, and using remote. All these actions are human-object interactions. The bounding box of the object in each frame is manually labelled.

\textbf{Sports-1M} \cite{karpathy2014large} contains 1, 133, 158 video URLs, which have been annotated automatically with 487 labels. It is one of the largest video datasets. Very diverse sports videos are included in this dataset, such as shaolin kung fu, wing chun, etc. The dataset is extremely challenging due to very large appearance and pose variations, significant camera motion, noisy background motion, etc. However, since the videos originate from Youtube and authors have not maintained the complete video set, some URLs become invalid or expired. Moreover, the automated labelling is known to have yielded incorrect groundtruth in some cases.

\textbf{Jester} \cite{materzynska2019jester} is a recent video dataset for hand gesture recognition. It is a large collection of densely-labeled video clips that shows humans performing pre-defined hand gestures in front of a laptop camera or webcam with a frame rate of 30 fps. There are in total 148,092 gesture videos under 27 classes performed by a large number of crowd workers. The dataset is divided into three subsets: training set (118,562 videos), validation set (14,787 videos), and test set (14,743 videos).

\textbf{OAD} \cite{li2016online} was captured with a Kinect camera \cite{zhang2012microsoft} in daily-life indoor environments and thus provides RGBD data. It includes 10 actions. The long video sequences in this dataset correspond to about 700 action instances. The starting and ending frames of each action are annotated in this dataset. 30 long sequences are used for training, and 20 long sequences are for testing.

\textbf{PKU-MMD} \cite{liu2017pku} is a large dataset for 3D activity analysis in continuous sequences. The data set contains 1076 long video sequences in 51 action categories, performed
by 66 subjects in three camera views. It contains almost 20,000 action instances and 5.4 million frames in total. This dataset also provides multi-modality data sources, including RGB, depth, Infrared Radiation and Skeleton.

\textbf{OoPS!} \cite{epstein2020oops} is proposed recently to study unintentional human actions. It includes a variety of daily human activities. Instead of focusing on action classification, the paper pays special attention to whether the performer succeeds/fails at achieving the goal. The dataset contains 20,338 videos and only has three class labels: Intentional, Transition and Unintentional. This paper conducted the action-intention prediction task on this dataset: early predicting the failure before it happens (1.5 seconds into future).

\pagebreak
\begin{landscape}
\vspace*{\fill}
\centering
\begin{longtable}[htbp]{@{}@{} l r r r r r r}
\toprule%
 \centering
 & \multicolumn{1}{c}{{{\bfseries Year}}}
 & \multicolumn{1}{c}{{{\bfseries No. Videos}}}
 & \multicolumn{1}{c}{{{\bfseries No. Actions}}}
 & \multicolumn{1}{c}{{{\bfseries Avg. Len}}}
 & \multicolumn{1}{c}{{{\bfseries Domain}}}
 & \multicolumn{1}{c}{{{\bfseries Data-Modality}}} \\

\cmidrule[0.4pt](l{0.25em}){1-1}%
\cmidrule[0.4pt](r{0.25em}){2-2}%
\cmidrule[0.4pt](r{0.25em}){3-3}%
\cmidrule[0.4pt](r{0.25em}){4-4}%
\cmidrule[0.4pt](r{0.25em}){5-5}%
\cmidrule[0.4pt](r{0.25em}){6-6}%
\cmidrule[0.4pt](r{0.25em}){7-7}%
\endhead

\textbf{OoPS!}  \cite{epstein2020oops} & 2020 &
20,338 & 3 & \highest{9.4s} & \highest{Action-intention} & RGB, Pose, Flow, Naaration \\ 

\myrowcolour
\textbf{Jester}  \cite{materzynska2019jester} & 2019 &
146,092 & 27 & \highest{3s} & \highest{Gesture} & RGB \\ 

\textbf{Something2something} \cite{goyal2017something} & 2017 & 
108,499 & 174 &  \highest{4.03} & \highest{Activity} & RGB \\

\myrowcolour
\textbf{PKU-MMD} \cite{liu2017pku} & 2017 &
1076 & 51 & \highest{-} & \highest{Activity, HOI} & RGB+D, IR, Skeleton\\ 

\textbf{OAD}  \cite{li2016online} & 2016 & 
59 & 10 & \highest{219} & \highest{Activity} & RGB+D, Skeleton\\

\myrowcolour
\textbf{NTU-RGBD} \cite{shahroudy2016ntu} & 2016 &
56,000 & 60 & \highest{-} & \highest{Activity, Interaction} & RGB+D, IR, 3D-Skeleton\\

\textbf{SYSU-3DHOI}  \cite{hu2015jointly} & 2015 &
480 & 12 & \highest{-} & \highest{Activity, HOI} & RGB+D \\

\myrowcolour
\textbf{UT-Interaction} \cite{ryoo2015ut} & 2015 & 
60 & 6 & \highest{3s} & \highest{Human Interaction} & RGB \\

\textbf{Sports-1M} \cite{karpathy2014large} & 2014 &
1,133,158 & 487 & \highest{-} & \highest{Action} & RGB \\

\myrowcolour
\textbf{ORGBD} \cite{yu2014discriminative} & 2014 & 
- & 7 & \highest{12} & \highest{Action, HOI} & RGB+D, Skeleton, BBx\\

\textbf{JHMDB-21}  \cite{jhuang2013towards} & 2013 & 
928 & 21 & \highest{-} & \highest{Action} & RGB, Flow, Skeleton, Contour \\

\myrowcolour
\textbf{UCF-101} \cite{soomro2012ucf101} & 2012 & 13,320 &
101 & \highest{6.39s} & \highest{Action} & RGB \\

\textbf{BIT} \cite{kong2012learning} & 2012 &
50 & 8 & \highest{-} & \highest{Human Interaction} & RGB \\

\myrowcolour
\textbf{HMDB-51} \cite{kuehne2011hmdb} & 2011 & 
6.766 & 51 & \highest{-} & \highest{Action} & RGB \\

\textbf{TV Human-Interaction} \cite{patron2010high} & 2010 &
300 & 20 & \highest{-} & \highest{Human Interaction} & RGB \\
\bottomrule
\caption[Summary of datasets for early action recognition.]{Summary of datasets for early action recognition.} \\
\end{longtable}
\vspace*{\fill}
\end{landscape}

{\bf Performance} evaluation of early action recognition has been focused on consideration of four of the above datasets: UCF-101, J-HMDB, UT-Interaction and BIT. We restrict performance comparisons to that set. For J-HMDB as shown in Table \ref{tab:jhmdb_rst}, we follow the standard procedure to report the recognition accuracy based on the first 20$\%$ frames of videos. Some work also reports results regarding the whole set of observation ratios and we refer readers to \cite{singh2017online, xu2019prediction} for more details. As noticed, recent advances 
based on motion extrapolation and Kalman filtering \cite{zhao2019spatiotemporal} as well as an RNN with RBF kernels \cite{shi2018action} have pushed the early recognition accuracy using only 20$\%$ frames to around $75\%$. Thus, there is still room for further improvements. Very recently, an effort achieved significant improvements (\ie $83\%$) via feature mapping learned through the Jaccard distance \cite{fernando2021anticipating}. 
On the other hand, for the UT-Interaction dataset, recent approaches can achieve over $90 \%$ accuracy with only the first $20\%$ frames and can obtain almost perfect accuracy over the first $50\%$; see Table \ref{tab:uti_rst}. For this dataset, a wide variety of approaches perform very well. These observations somewhat indicate that UT-Interaction is saturated. Similarly, the  UCF-101 dataset has been demonstrated to be easily recognizable using less than $30\%$ frames, as shown in Table \ref{tab:ucf_rst}. Again, a wide variety of aproaches are able to do well on this dataset. In contrast, the BIT dataset, for most approaches, still provides difficulties for early recognition, probably due to its limited size. Although, the most recent approaches can do very well having viewed $50\%$ or more of the data.
\begin{table}[htbp]
  \centering
  \caption[Summary of J-HMDB Performance]{Summary of J-HMDB Performance}\label{tab:jhmdb_rst}%
    \begin{tabular}{l|r}
    \toprule
    Methods & Accuracy @ 20\% \\
    \midrule
    \midrule
    Within-class Loss \cite{ma2016learning} & 33\% \\
    DP-SVM \cite{soomro2018online} & 5\% \\
    S-SVM \cite{soomro2018online} & 5\% \\
    Where/What \cite{soomro2016predicting} & 10\% \\
    Context-fusion \cite{jain2016recurrent} & 28\% \\
    ELSTM \cite{sadegh2017encouraging} & 55\% \\
    G. Singh's \cite{singh2017online} & 59\% \\
    TPnet \cite{singh2018predicting} & 60\% \\
    DR2N \cite{sun2019relational} & 66\% \\
    Pred-GAN \cite{xu2019prediction}& 67\% \\
    RBF-RNN \cite{shi2018action} & 73\% \\
    RGN-KF \cite{zhao2019spatiotemporal} & 78\% \\
    JVS+JCC+JFIP \cite{fernando2021anticipating} & 83.5\% \\
    \bottomrule
    \bottomrule
    \end{tabular}%
\end{table}%

\begin{table}[htbp]
  \centering
  \caption[Summary of UTI Performance.]{Summary of UTI Performance. Notice that some work reports results for set $1$ and set $2$ of the UTI dataset separately \eg \cite{chen2018part} , while others report the averaged results.}\label{tab:uti_rst}
    \begin{tabular}{l|r|r}
    \toprule
    Methods & Accuracy @ 20\% & Accuracy @ 50\% \\
    \midrule
    \midrule
    S-SVM \cite{soomro2018online} & 11.00 & 13.40 \\
    DP-SVM \cite{soomro2018online} & 13.00 & 14.60 \\
    CuboidBayes \cite{ryoo2011human} & 25.00 & 71.00 \\
    CuboidSVM \cite{ryoo2010overview} & 31.70 & 85.00 \\
    Context-fusion \cite{jain2016recurrent} & 45.00 & 65.00 \\
    Within-class Loss \cite{ma2016learning} & 48.00 & 60.00 \\
    IBoW \cite{ryoo2011human} & 65.00 & 81.70 \\
    DBoW \cite{ryoo2011human} & 70.00 & 85.00 \\
    BP-SVM \cite{laviers2009improving} & 65.00 & 83.30 \\
    ELSTM \cite{sadegh2017encouraging} & 84.00 & 90.00 \\
    Poselet \cite{raptis2013poselet} & - & 73.33 \\
    PA-DRL (set 1) \cite{chen2018part} & 69.00 & 91.70 \\
    PA-DRL (set 2) \cite{chen2018part} & 58.00 & 83.30 \\
    Future-dynamic Image & 89.20 & 91.90 \\
    Early-GAN \cite{wang2019early} & 38.00 & 78.00 \\
    Pred-GAN (set 1) \cite{xu2019prediction} &    -   & 100.00 \\
    Pred-GAN (set 2) \cite{xu2019prediction} &    -   & 85.70 \\
    RBF-RNN \cite{shi2018action} &   -    & 97.00 \\
    Joint-GAN \cite{gammulle2019predicting} & 98.30 & 99.20 \\
    SPR-Net \cite{hou2020confidence} & - & 85.30 \\
    \bottomrule
    \bottomrule
    \end{tabular}%
\end{table}%

\pagebreak
\begin{landscape}
\vspace*{\fill}
\begin{table}[!ht]
    \caption[Summary of UCF-101 Performance.]{Summary of UCF-101 Performance. Notice that some work only reports accuracy at a few ratios (\ie 50\% and 100\%). The symbol $\approx$ means approximated number from the original plotted graph, rather than actual number, due to the unavailability of some results.}\label{tab:ucf_rst}%
    \setlength{\tabcolsep}{0.5em} 
    {\renewcommand{\arraystretch}{1.3}
    \begin{tabular*}{\linewidth}{|cc| @{\extracolsep{\fill}} c|c|c|c|c|c|c|c|c|c|}
    \toprule & \multicolumn{1}{r|}{Metrics} & \multicolumn{10}{c|}{Obervation Ratio} \\
    \cmidrule{3-12}    \multicolumn{1}{|l}{Methods} &       & 10\%  & 20\%  & 30\%  & 40\%  & 50\%  & 60\%  & 70\%  & 80\%  & 90\%  & 100\% \\
    \midrule
    \multicolumn{2}{|c|}{IBoW \cite{ryoo2011human}} & 36.29 & 65.69 & 71.69 & 74.25 & 74.39 & 75.23 & 75.36 & 75.57 & 75.79 & 75.79 \\
    \multicolumn{2}{|c|}{DBoW \cite{ryoo2011human}} & 36.29 & 51.57 & 52.71 & 53.13 & 53.16 & 53.24 & 53.24 & 53.24 & 53.45 & 53.53 \\
    \multicolumn{2}{|c|}{MTSSVM \cite{kong2014discriminative}} & 40.05 & 72.83 & 80.02 & 82.18 & 82.39 & 83.21 & 83.37 & 83.51 & 83.69 & 82.82 \\
    \multicolumn{2}{|c|}{MSSC \cite{cao2013recognize}} & 34.05 & 53.31 & 58.55 & 57.94 & 61.79 & 60.86 & 63.17 & 63.64 & 61.63 & 61.63 \\
    \multicolumn{2}{|c|}{DeepSCN \cite{kong2017deep}} & 45.02 & 77.64 & 82.95 & 85.36 & 85.75 & 86.70 & 87.10 & 87.42 & 87.50 & 87.63 \\
    \multicolumn{2}{|c|}{Mem-LSTM \cite{kong2018action}} & 51.02 & 80.97 & 85.73 & 87.76 & 88.37 & 88.58 & 89.09 & 89.38 & 89.67 & 90.49 \\
    \multicolumn{2}{|c|}{MSRNN \cite{lan2014hierarchical}} & 68.00 & 87.39 & 88.16 & 88.79 & 89.24 & 89.67 & 89.85 & 90.28 & 90.43 & 90.70 \\
    \multicolumn{2}{|c|}{PA-DRL \cite{chen2018part}} & -     & -     & -     & -     & 87.30 & -     & -     & -     & -     & 87.70 \\
    \multicolumn{2}{|c|}{AAPNet \cite{kong2018adversarial}} & 59.85 & 80.85 & 86.78 & 86.47 & 86.94 & 88.34 & 88.34 & 89.85 & 90.85 & 91.99 \\
    \multicolumn{2}{|c|}{Student-Teacher \cite{wang2019progressive}} & 83.32 & 87.13 & 88.92 & 90.85 & 91.04 & 91.28 & 91.28 & 91.23 & 91.31 & 91.47 \\
    \multicolumn{2}{|c|}{Joint-GAN \cite{gammulle2019predicting}} & -     & 84.20     & -     & -     & 85.60 & -     & -     & -     & -     & - \\
    \multicolumn{2}{|c|}{RGN-KF \cite{zhao2019spatiotemporal}} & 83.12 & 85.16 & 88.44 & 90.78 & 91.42 & 92.03 & 92.00 & 93.19 & 93.13 & 93.13 \\
    \multicolumn{2}{|c|}{Pred-CGAN \cite{xu2019prediction}} & -     & -     & -     & -     & 91.14 & -     & -     & -     & -     & 93.76 \\
    \multicolumn{2}{|c|}{Eearly-CGAN $\approx$ \cite{wang2019early}} & 78.00 & 84.50 & 86.00 & 88.00 & 89.40 & 89.95 & 90.01 & 90.53 & 90.76 & 91.05 \\
    \multicolumn{2}{|c|}{SPR-Net \cite{hou2020confidence}} & 88.70 & - & - & - & 91.60 & - & - & - & - & 91.40 \\
    \multicolumn{2}{|c|}{STGCN \cite{fernando2021anticipating}} &80.26 &-& 89.86 &-& 92.87 &-& 94.08 &-& 94.43 &- \\
    \multicolumn{2}{|c|}{JVS+JCC+JFIP \cite{fernando2021anticipating}} & - & 91.70 & - & - & - & - & - & - & - & - \\
    \bottomrule
    \end{tabular*}%
}
\end{table}%
\vspace*{\fill}
\end{landscape}

\begin{landscape}
\vspace*{\fill}
\begin{table}[!ht]
    \caption{Summary of BIT-101 Performance}
    \setlength{\tabcolsep}{0.5em} 
    {\renewcommand{\arraystretch}{1.3}
    \noindent\begin{tabular*}{\linewidth}{|cc| @{\extracolsep{\fill}} c|c|c|c|c|c|c|c|c|c|}
        \toprule
          & \multicolumn{1}{r|}{Metrics} & \multicolumn{10}{c|}{Obervation Ratio} \\
\cmidrule{3-12}    \multicolumn{1}{|l}{Methods} &       & 10\%  & 20\%  & 30\%  & 40\%  & 50\%  & 60\%  & 70\%  & 80\%  & 90\%  & 100\% \\
    \midrule
    \multicolumn{2}{|c|}{IBoW \cite{ryoo2011human}} & 22.66 & 24.22 & 37.50 & 48.44 & 48.44 & 52.34 & 46.09 & 49.22 & 42.97 & 43.75 \\
    \multicolumn{2}{|c|}{DBoW \cite{ryoo2011human}} & 22.66 & 25.78 & 40.63 & 43.75 & 46.88 & 54.69 & 55.47 & 54.69 & 55.47 & 53.13 \\
    \multicolumn{2}{|c|}{MTSSVM \cite{kong2014discriminative}} & 28.12 & 32.81 & 45.31 & 55.47 & 60.00 & 61.72 & 67.19 & 70.31 & 71.09 & 76.56 \\
    \multicolumn{2}{|c|}{MSSC \cite{cao2013recognize}} & 21.09 & 25.00 & 41.41 & 43.75 & 48.44 & 57.03 & 60.16 & 62.50 & 66.40 & 67.97 \\
    \multicolumn{2}{|c|}{DeepSCN \cite{kong2017deep}} & 37.50 & 44.53 & 59.38 & 71.88 & 78.13 & 85.16 & 86.72 & 87.50 & 88.28 & 90.63 \\
    \multicolumn{2}{|c|}{MSDA \cite{chen2012marginalized}} & -     & -     & -     & -     & 70.00 & -     & -     & -     & -     & 81.50 \\
    \multicolumn{2}{|c|}{GLTSD \cite{lai2017global}} & 26.60     & -     & -     & -     & 79.40 & -     & -     & -     & -     & - \\
    \multicolumn{2}{|c|}{PA-DRL \cite{chen2018part}} & -     & -     & -     & -     & 85.90 & -     & -     & -     & -     & 91.40 \\
    \multicolumn{2}{|c|}{AAPNet \cite{kong2018adversarial}} & 38.84 & 45.31 & 64.84 & 73.40 & 80.47 & 88.28 & 88.28 & 89.06 & 89.84 & 91.40 \\
    \multicolumn{2}{|c|}{RGN-KF \cite{zhao2019spatiotemporal}} & 35.16 & 46.09 & 67.97 & 75.78 & 82.03 & 88.28 & 92.19 & 92.28 & 92.16 & 92.16 \\
    \multicolumn{2}{|c|}{Pred-CGAN \cite{xu2019prediction}} & -     & -     & -     & -     & 87.31 & -     & -     & -     & -     & 91.79 \\
    \multicolumn{2}{|c|}{Eearly-CGAN $\approx$ \cite{wang2019early}} & 38.80 & 58.00 & 79.00 & 90.00 & 96.00 & 96.00 & 96.00 & 96.00 & 96.00 & 96.00 \\
    \multicolumn{2}{|c|}{SPR-Net \cite{hou2020confidence}} & 84.10 & - & - & - & 100.00 & - & - & - & - & 99.20 \\
    \multicolumn{2}{|c|}{STGCN \cite{fernando2021anticipating}} & 46.09 & - & 58.59 & - & 81.25 & - & 89.06 & - & 86.72 & - \\
    \bottomrule
    \end{tabular*}%
    }
  \label{tab:bit_result}%
\end{table}%

\vspace*{\fill}
\end{landscape}
\pagebreak

\chapter{Future Action Prediction}\label{CH3}
\section{Overview}
\begin{figure}[htb]
    \centering
    \resizebox{.8\linewidth}{!}{\includegraphics{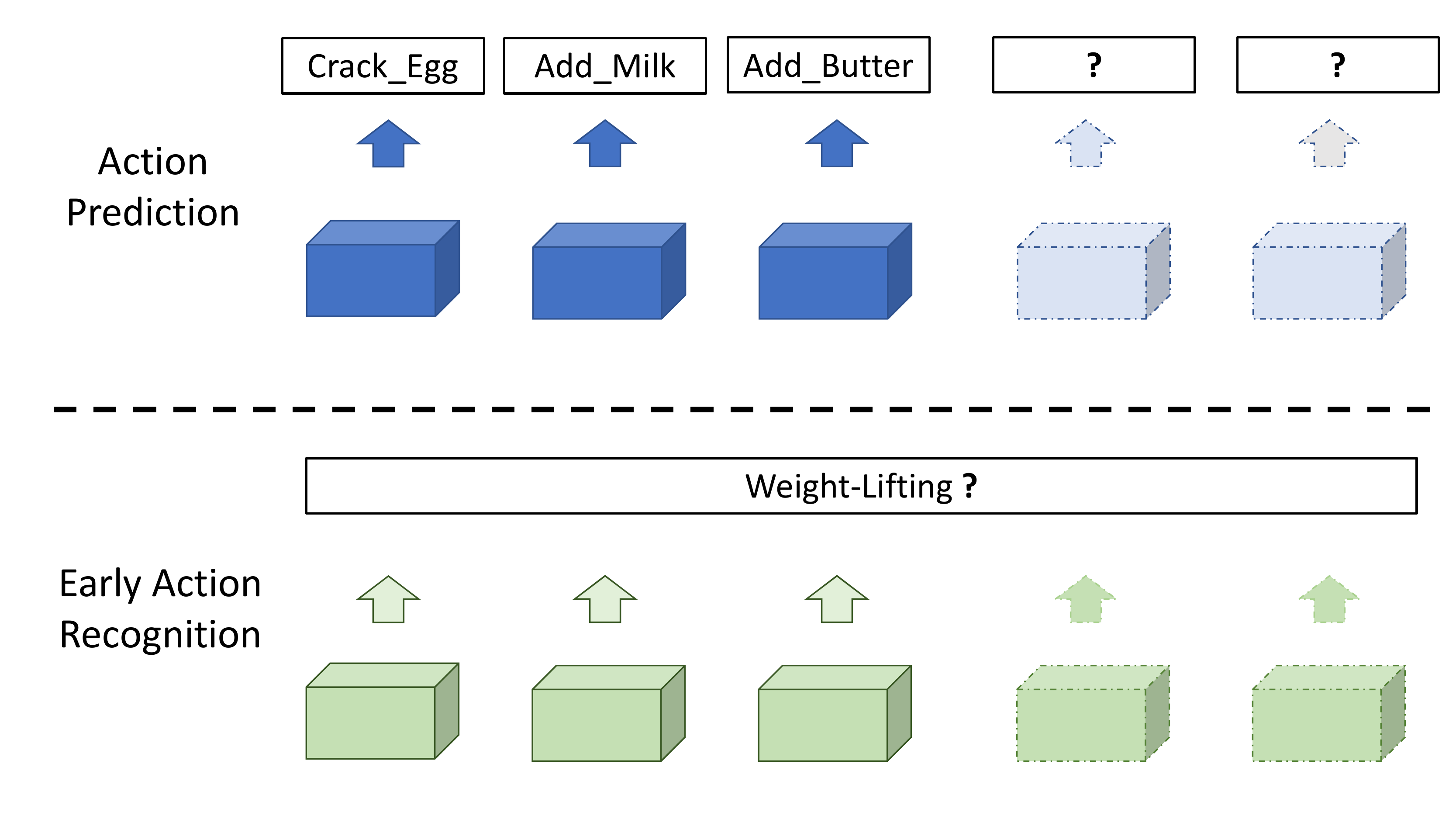}}
    \caption[Dissimilarity between future action prediction and early action recognition.]{Illustration of the general dissimilarity between future action prediction and early action recognition. The latter (bottom row) aims to infer a class label for an ongoing action from partial initial observations; the former aims to anticipate (strings of) future actions from initial observations.}
    \label{fig:action_vs_activity}
\end{figure}
Other than recognizing certain \textit{single} actions at its early stage, another emerging topic has caught researchers' attention, that is future action prediction. It differs from the previously discussed problem, early action recognition of Section \ref{Sec:early_actio}, in that its goal is to produce predictions of subsequent future actions. As illustrated in Figure \ref{fig:action_vs_activity}, the input (as solid boxes) and its output (as boundary-dotted boxes) in Early Recognition (bottom row) share the same label, such as \textit{\textbf{Weight-Lifting}}, whereas in Action Prediction (top row) input and output are often different yet logically related.

The ability to deduce possible future events from previous and current observations is one of the core functionalities needed by contemporary intelligent systems. In the example of automated traffic systems, \ie artifical integellience guided vehicles, every driver needs to be cautiously aware of surroundings and would benefit from smart prediction systems that can tell the specific next action of in-coming vehicles (\eg turning-right), to better plan its own motion in response (\eg slow-down or keep driving). Early action recognition does not satisfy such a demand, as it simply predicts the completion of an in progress action and thereby lacks sufficient prediction horizon to anticipate future actions. Exploring the plausible future is well studied in other fields, \eg weather forecasting and stock price prediction. In the computer vision community, however, researchers have only recently become heavily active in exploring solutions.

To examine future action prediction in detail, we group existing research into three major divisions: \textbf{1.} activity singleton or sequence prediction; \textbf{2.} joint prediction of activity semantics and times; \textbf{3.} ego-centric oriented action prediction. The first division discusses the most frequent setting for future action prediction: given the raw video frame observations, ${y_{1:t}}$, and/or its action semantics, $x_{1:t}$, produce the semantic label for actions that would happen afterward. The prediction can be either the immediate single following action, \ie $x_{t+1}$, or a sequence of future actions, \ie $x_{t+1: t+n}$.  
Note that the prediction provides only action semantics (\eg labels); time of occurrence and duration are left ambiguous. 
The second division considers not just prediction of sequences of future actions, but also their times. The last division tries to assist action forecastings in ego-centric settings, where observed frames are taken from first-person-view and thus mimic real life carry-on camera settings.

\section{Activity Singleton or Sequence Prediction}\label{Sec:future_action_1}
This section describes the general version of future action prediction, which is mapping the video input, that often contains a sequence of observed actions, into its subsequent sequence of actions, as shown in Figure \ref{fig:action_vs_activity} top row. The length of the predicted sequence can vary from one to many.

Hereby, we formally define the problem according to one of the initial works in future action prediction \cite{chakraborty2014context}. Let a sequence of observed video frames be given as $y^{obs} = (y_{1}, y_{2}, \ldots, y_{t} )$ and its corresponding frame-wise action labels be given as $\hat{x}^{obs} = ( x_{1}, x_{2}, \ldots, x_{t})$. 
Note that the symbol $x^{obs}$ represents the groundtruth video level labels annotated through human labor. However, in most studies, researchers would estimate the action labels with existing methods, rather than directly using the groundtruth video level labels. To clarify this nuance, we denote the estimated action labels as $\hat{x}^{obs}$.
Since actions can take large time spans, the frame-wise action labels can be either repeating or idle (no activity performed). Our task is to produce the future action semantics $x^{unobs} = (x_{t+1}, x_{t+2},...)$ that happen at unobserved time indices, \ie $t+n, n>0$. The set of future time indices of interest, $N=(n, n+1, n+2, \ldots)$, are referred to as the prediction horizon. So in other words, the task is to seek a function $f$ that maps $y^{obs} \rightarrow x^{unobs}$ with specified prediction horizon $n_{i}$. 

Indeed, compared with the standard video action recognition definition $f: y^{obs} \rightarrow x^{obs}$, the above definition seems like a temporal shifted version of the recognition function, \eg shifting $x^{obs}$ to $x^{unobs}$ by a desired time span. However, it is worth noticing that the semantic level information of observed video input, $y^{obs}$, can be inferred beforehand, either with groundtruth annotations or extraction algorithms, and therefore it is possible to operate a high-level semantic mapping $y^{obs} \rightarrow \hat{x}^{obs} \rightarrow x^{unobs}$ as a solution. Actually, most researchers adopted the second approach, especially for long-term predictions that typically cover multiple subsequent actions.

\begin{figure}[htb]
    \centering
    \resizebox{.7\linewidth}{!}{\includegraphics{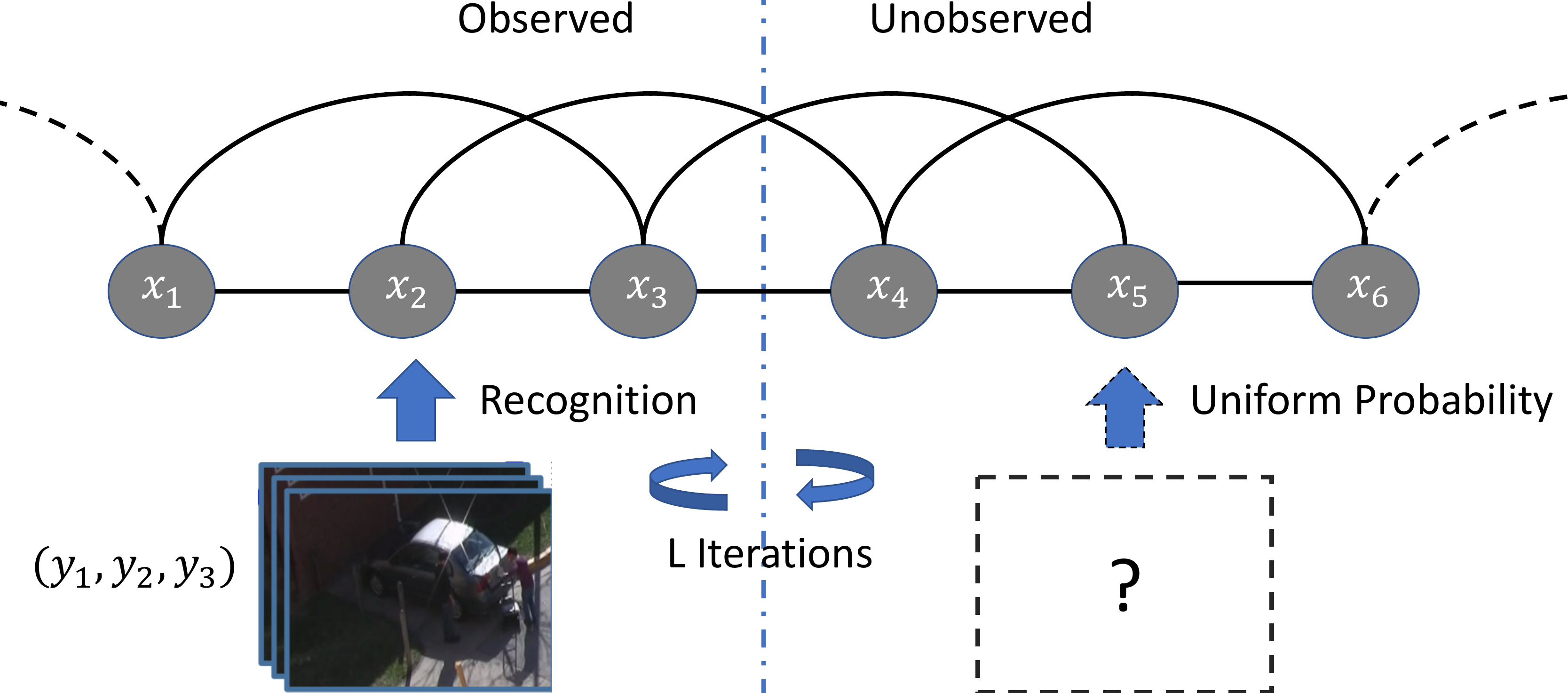}}
    \caption[MRF based approach for future action inference.]{Workflow diagram of MRF based approach from \cite{chakraborty2014context}. Some nodes are initialized as recognized activity labels from observed video frames (\eg $x_1, x_2, x_3$). Other nodes are initialized with uniform probability for entire action candidate set (\eg $x_4, x_5, x_6$). Loopy belief propagation (LBP) will run $L$ iterations to pass messages between nodes. As noted, the direct mapping from the past to the future is built upon high level action semantic labels $\hat{x}^{obs} \rightarrow x^{unobs}$, instead of raw frame inputs. The labels are extracted from raw frame inputs using a standard action recognition algorithm. Figure modified with permission from \cite{chakraborty2014context}}.
    \label{fig:accv2014_mrf}
\end{figure}

For example, early work along this line \cite{chakraborty2014context} proposed to treat any sequence of actions performed by all actors within a video as connected nodes in a Markov Random Field model (MRF) \cite{geman1986markov}, as shown in Figure \ref{fig:accv2014_mrf}. These nodes are connected with a manually defined edge potential function that reflects a reasonable range of spatiotemporal affinity between any two nodes. Furthermore, a node potential function that denotes how likely a certain observation, $y_{t}$, belongs to certain action label, $x_{t}$, was implemented as SVM classifiers.
More specifically, the observed action labels, $\hat{x}^{obs}$, come from a pre-trained recognition system that relies on a Bag-of-Word (BoW) approach over Space Time Interest Points (STIP) and Multi-class SVMs. To do the inference of the next action labels $x_{t+n}$, loopy belief propagation (also known as the sum-product algorithm) \cite{murphy2013loopy} is performed to compute the conditional distribution: $p(x^{unobs}| \hat{x}^{obs})$.
The capacity of a Markov model greatly depends on the order of dependency. The adopted MRF model in \cite{chakraborty2014context} specifies the maximum size of the neighboring clique as four nodes (two nodes in the past and two nodes in the future, if any), which leads to faster inference but limited performance. 

\begin{figure}[ht]
    \centering
    \begin{minipage}{.32\textwidth}
    \centering
    \includegraphics[width=1.\textwidth]{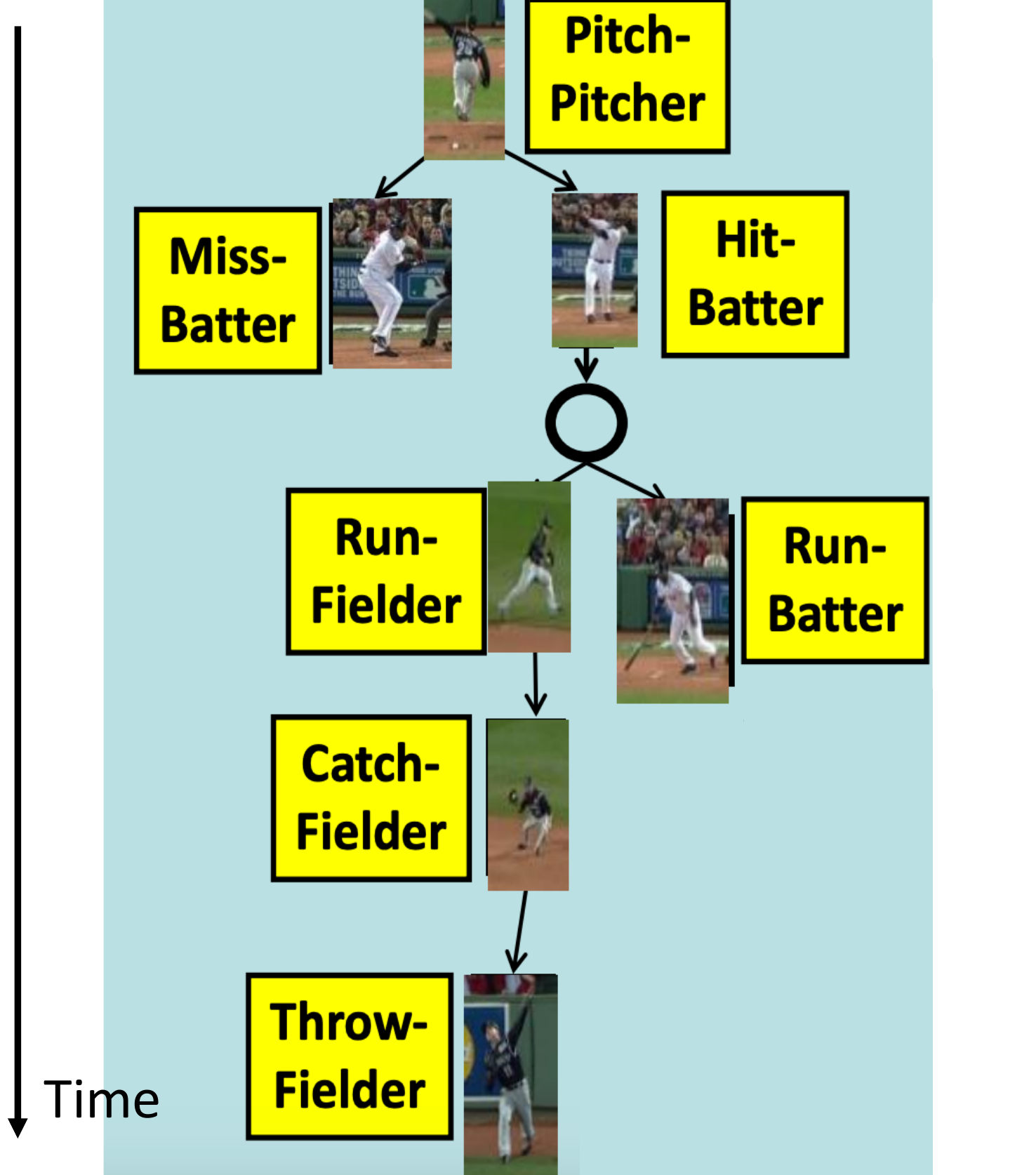}
    \caption*{(a)}
    \end{minipage}
    \begin{minipage}{.55\textwidth}
    \centering
    \includegraphics[width=1.\textwidth]{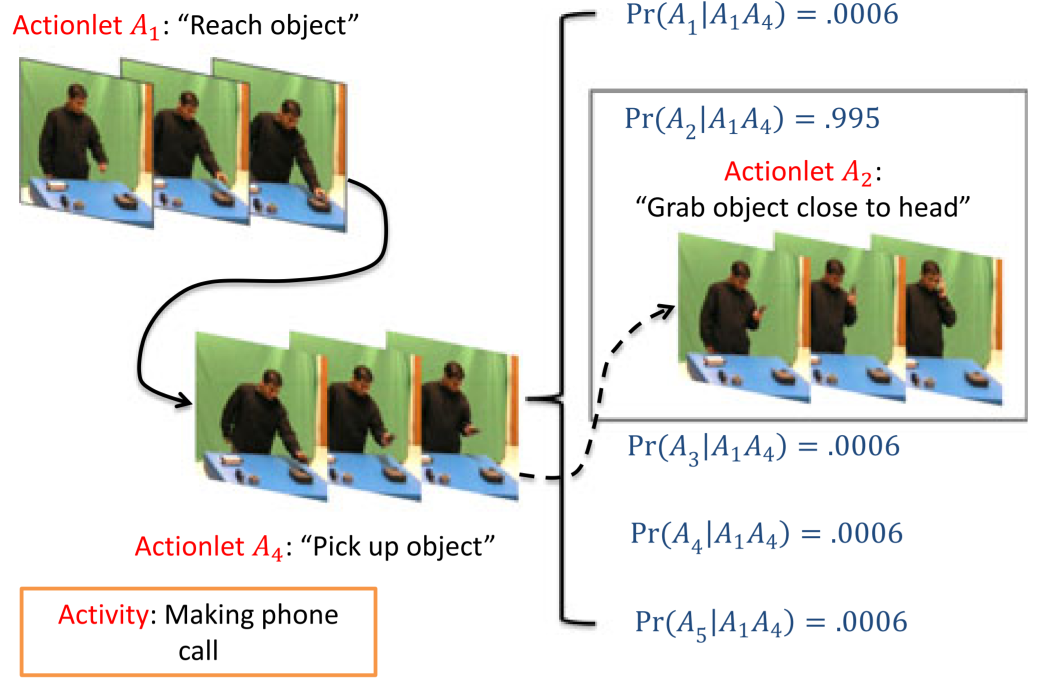}
    \caption*{(b)}
    \end{minipage}
\caption[Tree graph action prediction.]{(a) Example of representing an activity sequence as a tree graph from \cite{gupta2009understanding}. The temporal development is structured as tree nodes and leafs in a top-down fashion, where the top stands for the starting and bottom for ending. (b) The prediction of future action ``Grab object close to head" is selected by traversing the PST given prefix input and choosing the most likely suffix. Figure reproduced with permission from \cite{li2014prediction}.}
\label{fig:ngram_pst}
\end{figure}

Another research effort also adopted this high-level label inference scheme but relies on a more flexible Markov model, namely the Various order Markov dependency Model (VMM) implemented as graph (or tree) structures to model the temporal relation among activity sequences, as shown in Figure \ref{fig:ngram_pst} (a). Work that relies on such structures for future activity parsing include \cite{li2012modeling, pei2011parsing, li2014prediction, hamid2009novel}. These approaches advocate to represent activity sequence information as a sequence of ordered discrete alphabet symbols (grammars) and group them with \textit{n-gram} representations. Again, to obtain action labels from observed inputs, off-the-self action detection as well as recognition algorithms are needed.

As an example, some efforts choose to build a Probabilistic Suffix Tree (PST) for representing complex activity structures \cite{li2014prediction}. The PST structure is learned from training data, with the pretrained tree providing the chained transition probabilty for each suffix. Thus, the inference of future actions can be viewed as finding the most probable suffix leaf node given prefix node inputs, as seen in Figure \ref{fig:ngram_pst} right side.
The obvious drawback of Markov based methods is that the length of prediction is pre-defined in the training set. For example in \cite{li2014prediction}, the height of the PST is decided during training and thus the model can not be easily extended to undefined future lengths. However, the \textit{n-gram} representation of video activity is an over-simplified method, as critical information contained in image space is ignored. To this end, a recent effort researched on video representation learning in Riemannian geometry space, where the learned model can enable hyperbolic space embeddings, which is a continuous version of tree structures as shown in Figure \ref{fig:ngram_pst}. Thus, the predictive model supports building activity hierarchies for both early and future action prediction \cite{suris2021learning}.

Instead of using generative probabilistic models as above, a discriminative instance of Markov Chains, Conditional Random Fields (CRFs) \cite{lafferty2001conditional}, has been suggested to combine with particle filters for anticipating future human-object interactions \cite{koppula2015anticipating}. This combination enables a straightforward temporal extrapolation, as the particle filters are a type of recursive Bayesian filter that can propagate into unlimited time and space. In particular, they jointly modeled the future activity semantics with object affordances as well as movement trajectories. Notably, a recent work has advanced generative adversarial grammar learning for activity forecasting \cite{piergiovanni2020adversarial}.

The aforementioned approaches build models on a high-level semantic space, rather than raw video input or even visual features. The underlying assumption is that activity labels are linked by certain standard rules, \eg actions for making certain salads are dictated by their instructions. Therefore, straightforward modelling on sequences of action labels is sufficient. Yet in real life, scenarios exist where future actions are less well predicted by well defined scripts. For example, pedestrians walking can happen in a wide range of places and times. It can lead to ``road crossing", ``turning", ``stop and watch", etc. So, reasoning the future from an extremely common action is hard. Furthermore, in surveillance camera settings, actors can enter the field of view at times that obscure their previous actions, in which cases observing one's full activity sequence is impossible. To solve such problems, many researchers resort to a direct mapping from $y^{obs}$ to future actions $x^{unobs}$.
\begin{figure}[htb]
    \centering
    \resizebox{1.\linewidth}{!}{\includegraphics{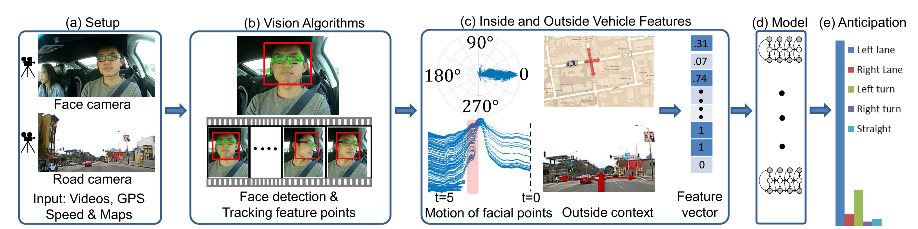}}
    \caption[Multi-sensors next action anticipation.]{The setting of \cite{jain2015car} combing multiple cues. Multiple cues collected from both inside and outside sensors are used to anticipate the next maneuver of drivers. Figure reproduced with permission from \cite{jain2015car}.}
    \label{fig:aio_hmm}
\end{figure}

One early effort that followed this scheme proposed to rely on cues distilled from multiple visual sources to predict the next driver maneuver \cite{jain2015car}. Specifically, it collected information from four information sources: 1. facial motion trajectories and head angles detected from the driver-facing camera and feature tracking techniques; 2. speed logger from the vehicle system; 3. outside context from the road-facing camera; 4. the GPS map. They fused these four types of features with a novel HMM model where the maneuvers of drivers play as the hidden variables, outside sensory as control units and inside sensory as observations. However, since HMMs can not handle very high dimensional data, they converted most sensory cues into binary features, \eg if vehicle speed is larger than 15km/hr, set the feature flag as 1 otherwise 0.

To encourage more expressive features, \eg 3D human head/facial models, the authors made an upgrade on the sequential modelling module \cite{jain2016recurrent}. They replaced the HMMs with a recurrent neural network (an LSTM) and modulated four sensory channels with two individual RNNs. The final fusion was done by MLPs, followed by softmax classifiers. A critical difference between this work and earlier related approaches based on high-level intermediate inferences, e.g., \cite{li2014prediction}, is the encompassed time span. Observations in the high-level based usually cover multiple actions that can takes \textbf{minutes}. In contrast, \cite{jain2016recurrent} only consumes frames that are 0.8 second (about 20 frames) ahead of the event happening, which has the potential to be deployed in real-time systems.

Similar work that focuses on short-term (or singleton) action prediction directly from past raw video inputs appears frequently in many sub-areas. One such effort proposed to leverage environment-agent interactions to anticipate risk actions as well as localize future risk regions \cite{zeng2017agent}. Another worked on anticipating traffic accidents from vehicle cameras through stacking local detected object features and global context \cite{suzuki2018anticipating}. Some other work built a graph Convolutional Network (GCN) to model interactions between multiple agents and contextual objects to forecast future activities along with future trajectories of pedestrians \cite{liang2019peeking}. Pedestrian road-crossing intention prediction is one concrete example of action prediction in traffic scenarios \cite{rasouli2019pie, rasouli2020pedestrian, rasouli2017they}. Again, in these works, past observations contain little or no evidently useful activity semantics, instead researchers found it helpful to mine cues from subtle details.
As an example, an effort on anticipating crossing of pedestrians fused human poses, human visual representation extracted from deep networks, scene contextual information and vehicle speed, with a RNN, to make predictions \cite{rasouli2020pedestrian}.

\section{Joint Prediction of Activity and Time}
\begin{figure}[htb]
    \centering
    \resizebox{.8\linewidth}{!}{\includegraphics{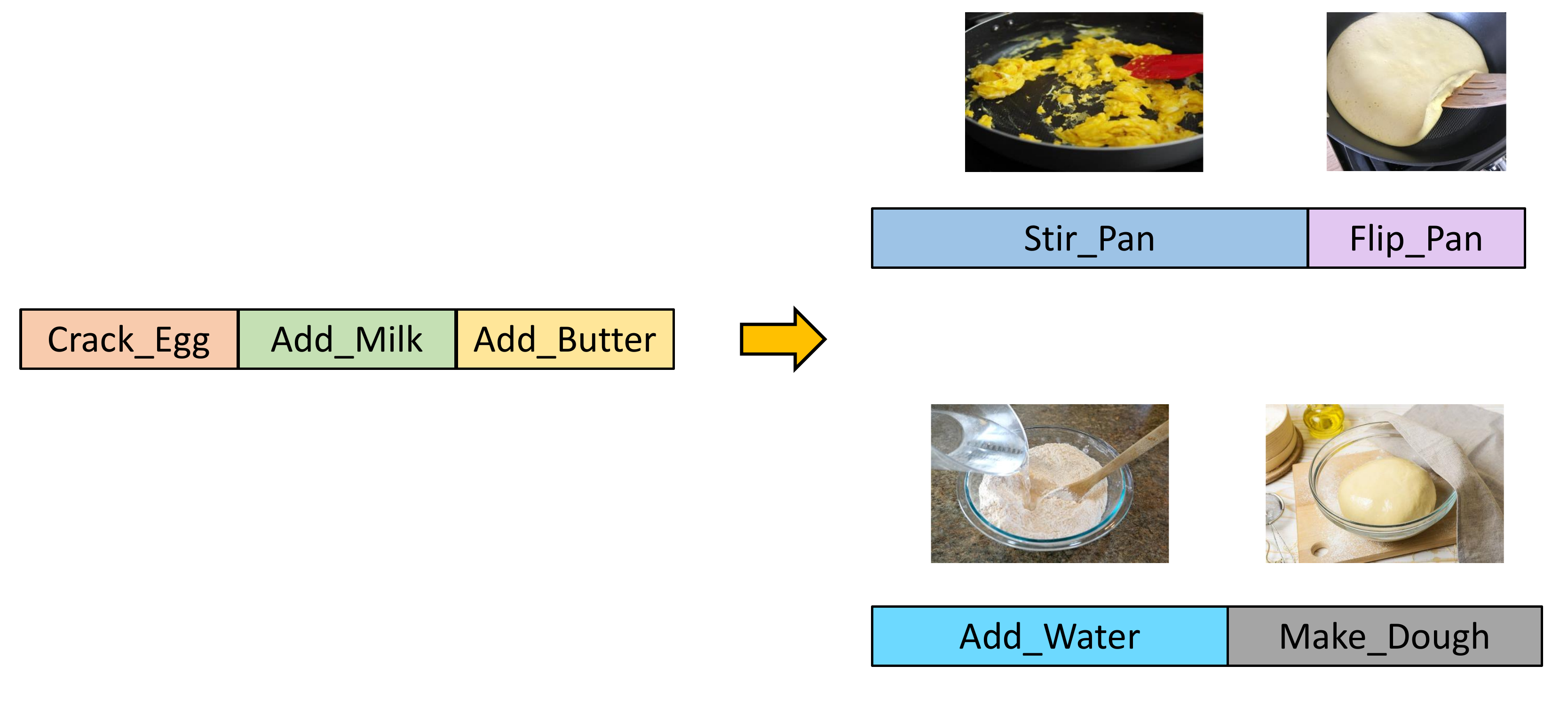}}
    \caption[Predicting future activity labels and time durations.]{Predicting future activity labels and time durations. The left hand side shows initial observed activities, indicated as labelled boxes, with their temporal duration encoded by box length. The right hand side shows possible predictions.}
    \label{fig:act_time}
\end{figure}

Joint prediction of future actions and their times has recently become an active research area for its importance in critical applications like automated navigation, human-computer interaction as well as for the intellectual challenges it presents. All video-based action prediction work forecasts upcoming activities, here, the distinction is that both the activities and their times are predicted. An example is provided in Figure \ref{fig:act_time}.

\begin{figure}
    \centering
    \begin{minipage}{.4\textwidth}
    \centering
    \includegraphics[width=1.\textwidth]{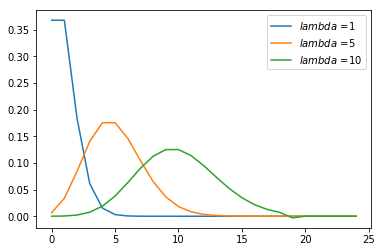}
    \caption*{(a)}
    \end{minipage}
    \begin{minipage}{.4\textwidth}
    \centering
    \includegraphics[width=1.\textwidth]{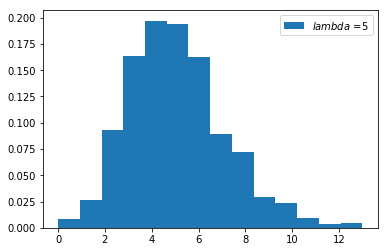}
    \caption*{(b)}
    \end{minipage}
\caption[Poisson distribution.]{(a) Probability density function of Poisson distribution regarding various parameters $\lambda$. (b) Histogram distribution of 1000 sampled inter-arrival times from Poisson process with $\lambda = 5$. Since Poisson is a member of the exponential family, its distribution looks similar to a Gaussian distribution, but its data points are always positive integers.}
\label{fig:poisson_intuition}
\end{figure}

An initial effort in this direction \cite{mahmud2016poisson} adopted the Poisson process \cite{kingman2005p} as a key technique for activity inter-arrival time modelling. Notice that inter-arrival time differs from the absolute activity occurrence time in that it stands for the duration between the starting and ending of a certain action. The probability density function for the possible next action interval, $\tau$, under the assumed Poisson distribution is
\begin{equation}
    p(T = \tau) = \frac{\lambda^{\tau}}{\tau\,!} e^{\lambda} 
\label{eq:possion}
\end{equation}
where $\lambda$ is the intensity function that controls the shape of the Poisson probability density function and its physical meaning is the average number of events among the whole time span. The Poisson distribution and sampling are plotted at Figure \ref{fig:poisson_intuition} (a) and (b) respectively.

There are three properties of the Poisson distribution that are worth noticing: \textbf{1.} Its samples are always positive integers. Equation \ref{eq:possion} incurs a \textbf{factorial} function on time $\tau$, thus its result must be an integer; additionally, by default researchers assume there is no negative time value, so that $\tau > 0$. To summarize, the Poisson distribution is a positive \textbf{discrete} function as shown in Figure \ref{fig:poisson_intuition} (b). \textbf{2.} Its shape is unimodal. \textbf{3.} It is memoryless in that the sampling of the current step is independent from any previous samples.

To support a stochastic modelling of time, the authors pre-defined a Gaussian Process prior for $\lambda(t)$ and learned its parameters (\ie the mean and variance for certain actions and times) from training data for every action. Then, they performed testing through importance sampling. Unfortunately, they only train and test on a single action input and single action output, without considering recursive temporal predictions.

\begin{figure}[htb]
    \centering
    \resizebox{1.\linewidth}{!}{\includegraphics{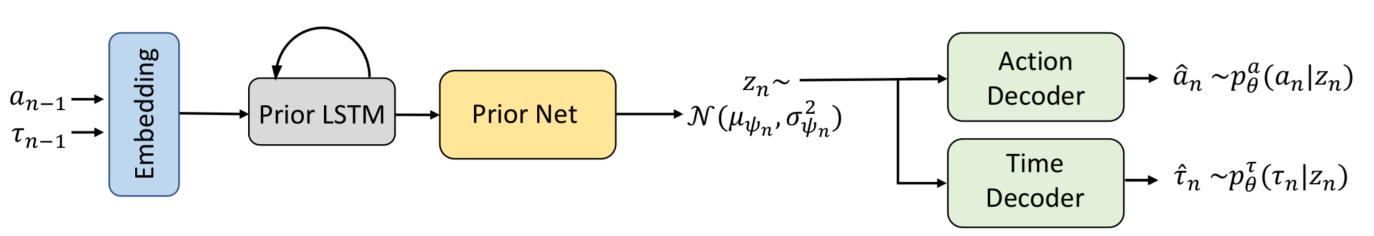}}
    \caption[VAE-Poisson Process.]{Predicting the next-step action and interval time in \cite{mehrasa2019variational}. Past action and time pairs $(a_{n-1}, \tau_{n-1})$ are used to stochastically produce the future action-time pair from sampled latent variable $z_{n}$ as well as the Action/Time Decoders. Figure reproduced with permission from \cite{mehrasa2019variational}}.
    \label{fig:vae_app}
\end{figure}

An extension of the initial work \cite{mahmud2016poisson} used the temporal point process for the same task \cite{mehrasa2019variational}, where the intensity parameter $\lambda$ is dynamically calculated from past observations as
\begin{equation}
    p(T = \tau) = \lambda(\tau | x_{1:k}) e^{-\int^{t|x_{1:k}}_{0} \lambda(t) dt},
\label{eq:temporal_possion}
\end{equation}
where $x_{1:k}$ represents past information, such as raw frames, semantic labels or past action times. 
Their model also supports stochastic time and action prediction via sampling the parameter $\lambda$ through a Variational Auto-Encoder (VAE) embodied as LSTMs. This approach is depicted in Figure \ref{fig:vae_app}. Due to the nature of Poisson distribution, this approach can not produce more than one future step (action, time) at each time step. Notably, both efforts \cite{mahmud2016poisson, mehrasa2019variational} resort to a learning fashion for the hyper-parameter $\lambda$.
\begin{figure}[htb]
    \centering
    \resizebox{.5\linewidth}{!}{\includegraphics{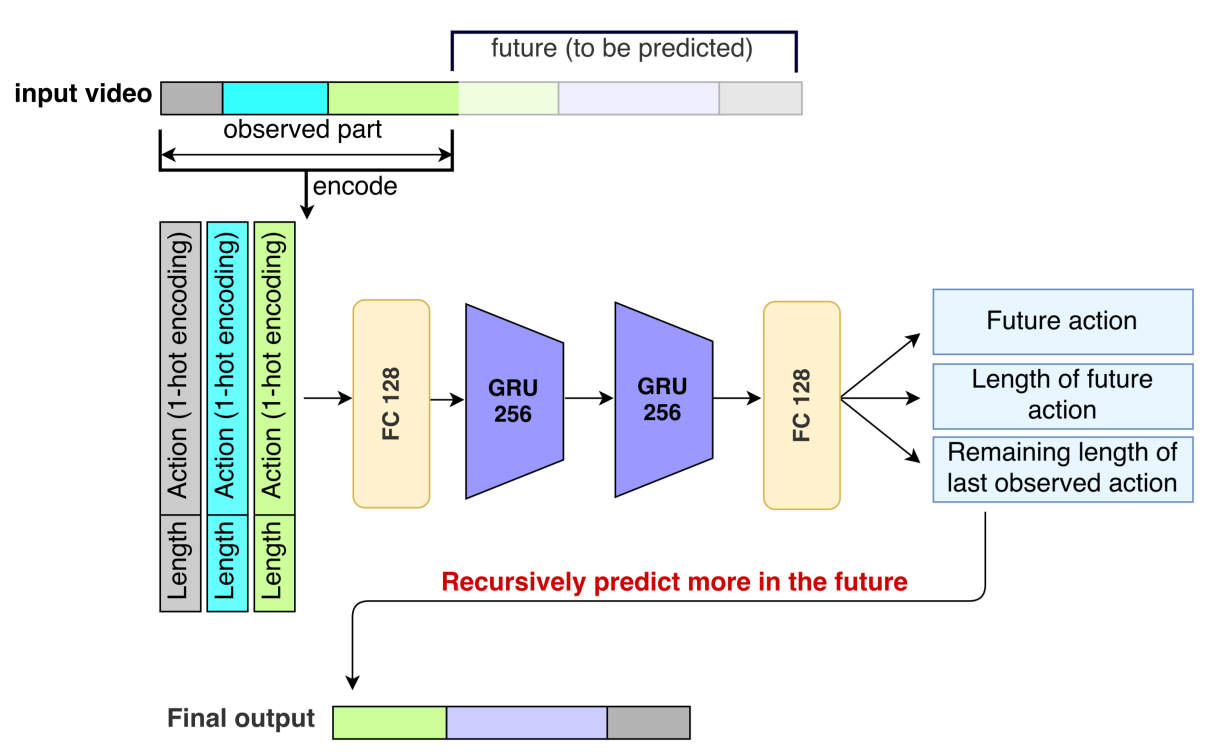}}
    \caption[Multi-task model for joint prediction of action and time.]{Model introduced in \cite{abu2018will} for action and time prediction via regression. Notice the observed input action is pre-processed into a one-hot encoding and the input time is framed as a real value scalar. Subsequent processing alternates use of fully connected (FC) layers and Gated Recurrent Units (GRUs). The whole scope of the future requires the recursive inference. Figure reproduced with permission from \cite{abu2018will}.}
    \label{fig:when_will}
\end{figure}

Though the Poisson process has been known for modelling time arrival events, not all researchers adopt it. An alternative research direction modeled time as a real-valued output directly from regression models, e.g., using an MLP to generate the output \cite{mahmud2017joint, abu2018will, abu2019uncertainty, ke2019time, gammulle2019forecasting} as shown in Figure \ref{fig:when_will}: For any next-step action, the ``length of future action" and ``remaining length of current action" will be regressed from fully-connected layers. In comparison to previous Poisson methods \cite{mehrasa2019variational, mahmud2016poisson}, these approaches choose deterministic time models rather than the Poisson distribution and therefore their outputs for a given input will be fixed. An exception to this general trend instead assumes a Gaussian prior and approximates the mean and variance via learning \cite{abu2019uncertainty}.

A third way to model temporal data is to treat time in terms of discrete categorical values. Along these lines, research on recommendation systems found that modelling time with a categorical representation and optimizing with cross-entropy consistently outperformed the use of real valued modelling \cite{li2017time}.
It is worth noticing that the range of the discrete categorical variable normally needs to be defined beforehand. For example, in image classifications, the ImageNet dataset defines 1000 categories and thus its models (\ie AlexNet) would output probability score vectors $\in \mathcal{R}^{1000}$. For discrete times, the maximum time index is often chosen as the time range value \cite{neumann2019future}.

To clarify the variability of the design choice for time representation, especially in video understanding tasks, a systematic study of the aforementioned choices was conducted \cite{neumann2019future}. This study grouped time representations into three categories, continuous, discrete and hybrid; see Table \ref{tab:time_representation}.
\begin{table}[htb]
\scriptsize
\centering
\begin{tabular}{|c|l|l|}
\hline
\multicolumn{1}{|c|}{Group} &       \multicolumn{1}{c|}{Method}                     & \multicolumn{1}{c|}{Detail}                                                               \\ \hline
\multirow{3}{*}{Discrete}   & One-in-many Classification & Quantize time in $\Delta{max}$ discrete bins and use classified index as results.                         \\ \cline{2-3} 
                            & Binary Classifiers         & $\Delta{max}$ independent binary classifiers for each time index.                                         \\ \cline{2-3} 
                            & Heuristic Heat-Map         & Generate 1d heat-map and choose the maximum position in heatmap.                            \\ \hline
\multirow{3}{*}{Continuous} & Direct Regression          & Real value regression.                                                                      \\ \cline{2-3} 
                            & Gaussian Distribution      & Estimate time as parameterized Normal $N(t | \mu, \sigma)$ distribution.                    \\ \cline{2-3} 
                            & Weibull Distribution       & Estimate tme as parameterized Weibull $p(t | \alpha, \beta) = 1 - e^{-(t/\alpha)^{\beta}}$. \\ \hline
Hybrid                      & Gaussian Mixture        & Gaussian Mixture of heat-maps. A hybrid model.                                              \\ \hline
\end{tabular}
\caption[Summary of time representations.]{Summary of time representations. $\Delta_{max}$ stands for the maximum time range in datasets.}
\label{tab:time_representation}
\end{table}

\begin{table}[htbp]
  \centering
    \begin{tabular}{lccc}
    \toprule
    Model & EPA[\%] & TTEE[s] & MS \\
    \midrule
    \midrule
    One-in-many Classifier & 71.78 & 2.66  & N/A \\
    Binary Classifier & 75.90  & 3.21  & N/A \\
    Direct Regression & 71.11 & 3.81  & N/A \\
    Gaussian Distribution & 66.77 & 5.77  & 10.91 \\
    Weibull Distribution & 55.66 & 8.31  & 3.20 \\
    Heuristic Heatmap & 76.15 & 2.64  & 3.88 \\
    \midrule
    Gaussian Mixture of Heuristic Heatmap & 78.70  & 2.04  & 2.69 \\
    \bottomrule
    \bottomrule
    \end{tabular}%
  \label{tab:bdd_time}%
  \caption[Comparison results of various time modelling methods.]{Comparison results of various time methods on BDD100K dataset. For evaluation metrics, higher Event Prediction Accuracy (EPA) means better performance, whereas lower Time-to-Event Error (TTEE) and Model Surprise (MS) mean better performance. Table reproduced with permission from \cite{neumann2019future}.}
    \label{fig:time_compare}
\end{table}%

In empirical comparison, there were three discrete methods (One-in-many Classifier, Binary Classifier and Heuristic Heatmap), three continuous methods (Direct Regression, Gaussian Distribution and Weibull Distribution) and one hybrid method (Gaussian Mixture of Heuristic Heatmap). These time representations are evaluated on the BDD100K car stop anticipation dataset \cite{yu2020bdd100k} that measures the Event Prediction Accuracy (EPA), Time-To-Event-Error (TTEE) and Model Surprise (MS). In their results (shown in Table \ref{fig:time_compare}), almost all continuous time modelling methods fall short of achieving comparable performance with discrete methods. Moreover, the hybrid model performs the best. This observation on video understanding agrees with the experimental results from other related fields \cite{li2017time}.

\begin{figure}[htb]
    \centering
    \resizebox{.9\linewidth}{!}{\includegraphics{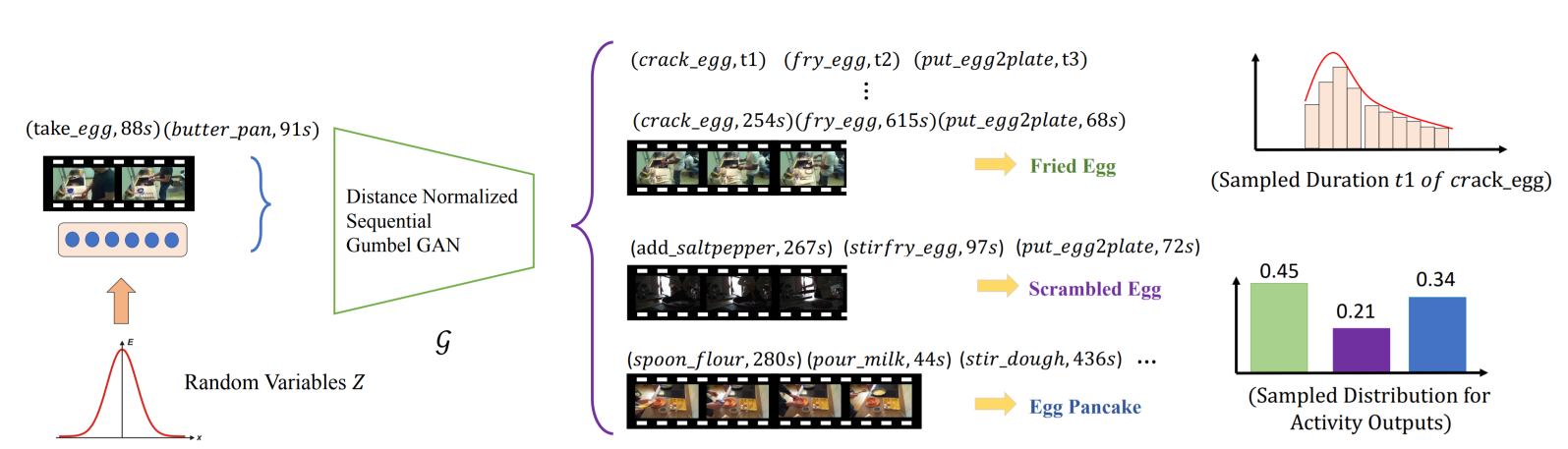}}
    \caption[Diverse action and time prediction.]{Method introduced in \cite{zhaodiverse} that used a sequential modeller (Distance Normalized Gumbel GAN) to produce the whole sequence of subsequent actions and times. Multi-modality is well supported for both action categories and time durations. Figure reproduced with permission from \cite{zhaodiverse}.}
    \label{fig:zhao_eccv2020}
\end{figure}

One possible explanation to validate the above phenomenon could be the effect of temporal scope prior, $\Delta_{max}$, enforced by discrete models. Continuous methods generate outputs in the real positive number realm, $\mathcal{R}^{+}$, while discrete methods pre-confine the output range to $\mathcal{R}^{\lambda \Delta_{max}}$ ($\lambda$ a scaling factor). This prior implicitly informs the learned prediction model about the reasonable temporal output range, whereas the continuous counterpart tries to cover the whole positive value realm. 
Another potential reason for the observed pattern of results could be the time data format. Most such work (\eg \cite{abu2018will, mehrasa2019variational, ke2019time}) assumes the minimum unit of time is in the unit of seconds, which corresponds to the data formatting common in adopted datasets (\eg Breakfast dataset \cite{kuehne2016end}). Discrete time format matches the original scheme naturally while continuous ones incur unnecessary decimals. Recent work succeeds the discrete time scheme and trains a conditional GAN model based on Gumbel discrete sampling to jointly enhance the accuracy and diversity in both future action semantics and times \cite{zhaodiverse}; see Figure \ref{fig:zhao_eccv2020}. Previous efforts tended to be challenged in producing realistic (accurate) and diverse predictions.

In summary, approaches to joint prediction of actions and their times can be categorized along three dimensions, as follows.

\noindent\textbf{Difference in Inputs:} Similar to the discussion on input data use in Section \ref{Sec:future_action_1}, research in time prediction also can be categorized according to the input data type. Some work aims at learning the mapping from high-level information, extracted from raw observations, to the future. These efforts usually depend on extracted/estimated activity labels and segmented temporal duration as input \cite{abu2018will, abu2019uncertainty, mehrasa2019variational, ke2019time, zhaodiverse}. Other work relies on raw video inputs \cite{neumann2019future, mahmud2017joint}.

\noindent\textbf{Difference in stochastic process:} The majority of work relies on a stochastic process for future actions and times. Various approaches employ a Poisson \cite{mahmud2016poisson,mehrasa2019variational} or Gaussian distributions \cite{abu2019uncertainty} or a generative adversarial network \cite{zhaodiverse}. A few outliers deterministically produce their anticipations \cite{abu2018will, li2017time}.

\noindent\textbf{Difference in prediction horizon:} Some work advocates for one-step future predictions. RNNs have been used to recursively generate next step (action, time) pairs (\eg  \cite{abu2018will}). To obtain the whole scope of the future, results of next-step prediction are reused as input for the next-next. This procedure repeats until it reaches the end of sequence (\eg \cite{mehrasa2019variational, gammulle2019forecasting, mahmud2017joint, neumann2019future}). Alternatively, other approaches generate the rest of the sequence in one shot (\eg \cite{zhaodiverse, abu2019uncertainty}).
Finally, work has made one-shot time independent predictions, so that predictions on remote time indices do not rely on previous ones \cite{ke2019time}. They achieve this goal by making predictions conditioned on the desired prediction index as an extra input. The evident downside of their work is the limited prediction scope.


\pagebreak

\section{Ego-Centric Action Prediction}
\begin{figure}[htb]
    \centering
    \resizebox{.9\linewidth}{!}{\includegraphics{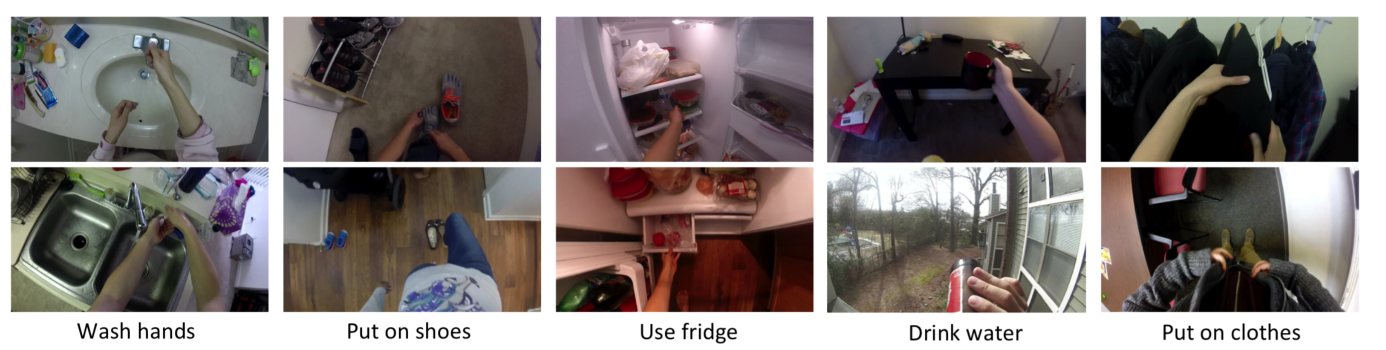}}
    \caption{Examples of Ego-centric activity footage.}
    \label{fig:ego_centric}
\end{figure}

First person video analysis has attracted increasing attention due to the rich information contained in egocentric visual data and easier access to wearable recording devices. As seen in Figure \ref{fig:ego_centric}, ego-centric filming exhibits much closer viewing angles of objects, environments and interactions of daily activity. However, one critical difference from the third person view dataset (\eg UCF-101 \cite{soomro2012ucf101}) is the lack of notable motion of actors, \ie only the hand motion is typically visible. Moreover, since the camera moves along with its carrier, there usually exists large background motion as well as jittering in video frames.

Missing actors in activity videos makes classic human action analysis methodologies not naturally applicable and larger view-point changes as well as camera jittering makes motion feature extraction (\eg optical flow) very noisy. These obstacles incentivise researchers to tackle ego-centric data in a slightly different way. 

In the following, we group related research work into two main divisions: \textbf{1.} work that actively explores non-traditional visual cues (\ie hand motion, gaze salience, object trajectory, etc.) that are uniquely salient in ego-view data; \textbf{2.} work that follows regular techniques and treats ego-centric data in a similar way as third-person view data. In this approach, even though videos are filmed under ego-centric view, performed actions can be inferred without direct observation of large portions of the actor.
We first discuss the former division, unique features, from the perspectives of gaze, hand and object.

\noindent\textbf{Gaze Information:} One beneficial information source hidden in first person video is gaze information. Gaze moving on/off of a certain object can closely correspond to the occurrence/ending of a particular action. Much work has enjoyed the incorporation of such information (\eg \cite{su2017predicting, lepetit2009epnp, shen2018egocentric}). 
 
\begin{figure}[ht]
    \centering
    \begin{minipage}{.3\textwidth}
    \centering
    \includegraphics[width=1.\textwidth]{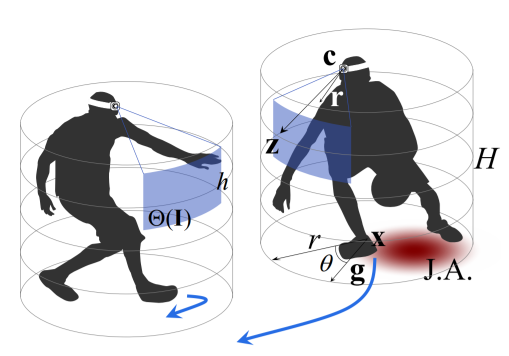}
    \caption*{(a)}
    \end{minipage}
    \begin{minipage}{.25\textwidth}
    \centering
    \includegraphics[width=1.\textwidth]{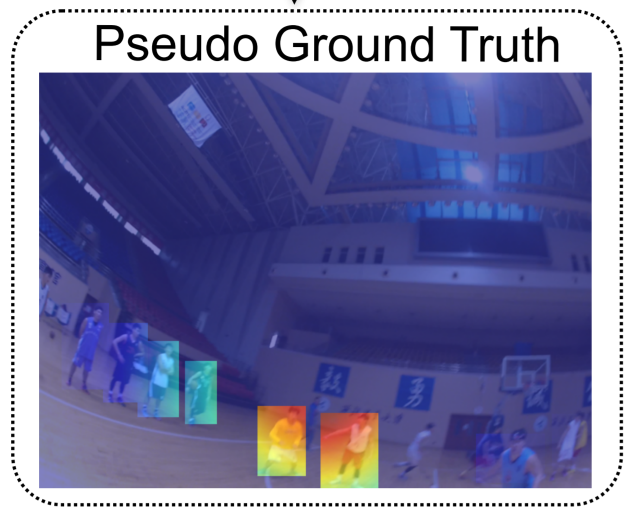}
    \caption*{(b)}
    \end{minipage}
    \begin{minipage}{.4\textwidth}
    \centering
    \includegraphics[width=1.\textwidth]{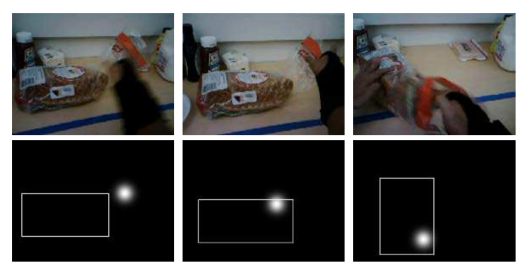}
    \caption*{(c)}
    \end{minipage}
    \caption[Examples of ego-centric methods.]{(a) Examples of 3D reconstruction of stablized basketball players and the recovered gaze direction $\mathbf{z}$ from \cite{su2017predicting}. (b) Example of next player intention groundtruth generated from ego-supervised prior knowledge with high probability score denoted as red, while low score is blue. (c) Gaze-events defined from \cite{shen2018egocentric} where object target detection (white rectangles) \textbf{overlaps} with gaze fixations (white circle). Subfigures (a), (b) and (c) reproduced with permission from \cite{su2017predicting}, \cite{bertasius2017using} and \cite{shen2018egocentric}, respectively.}
\label{fig:gaze}
\end{figure}

Yet, gaze localization does not come freely. One way to obtain gaze information is the 3D reconstruction of actors and then inferring the gaze direction from head direction and body poses \cite{su2017predicting}. In that work, the authors virtually stabilized input first person video and reconstructed 3D body pose by applying cylindrical projection using existing tools \cite{lepetit2009epnp}. This processing provided labels for the location, orientation, and velocity of actors with pixel level precision, as seen in Figure \ref{fig:gaze} (a). The inferred gaze direction and body pose together indicate the potential social interactions among players and thus contribute to anticipating future behaviors. 

Another way to gain gaze information is through manual annotation and/or use of additional instrumentation. Though expensive, it is easily doable through eye-tracking devices. Assuming decent gaze estimation is given, an effort proposed a model to focus on video frames where target objects and gaze estimations coincide \cite{shen2018egocentric} (shown in Figure \ref{fig:gaze} (c)). They paid special attention to gaze-events and distilled attention values from event-fired frames to assist action prediction.  

Some other work that considers gaze as an important visual cue focused efforts on predicting future gaze fixations from observed frames \cite{zhang2017deep}. Their framework proceeded in a similar way to video frame prediction but output binary gaze heat-maps. However, the authors did not show any application of the predicted gaze on other related problems.

\noindent\textbf{Hand Information:} In first person videos, hand movement typically is the major human visible action. Human-object interactions often consist of a movement part and one or more object parts, with “take bowl” as an example. First person videos capture complex hand movements during a rich set of interactions, thus providing a powerful workhorse for studying the connection between hand actions and future representation.

In imitation of the acquisition of gaze information, hand motion can be either annotated or online estimated. For example, an extant hand shape detector \cite{zhu2014pixel} can be applied to obtain a hand mask across frames \cite{shen2018egocentric}. The use of the estimated hand mask is straightforward: Overlaying the hand mask on RGB input frames.
In other recent work, it is proposed to use the hand movement trajectory as an output regularizer \cite{liu2019forecasting}. The authors enforced the learned network model to jointly predict future actions, future hand trajectories and interactive hotspots (regions where interaction would happen with high likelihood). In an ablation study, it was shown that the three related tasks were complementary to each other.

\noindent\textbf{Object Information:} 
Objects that are interacted with are another potentially important source of information for predicting future actions. While such information can be generally useful for action prediction, it can be of particular importance in first-person video where the movements of the actor may be obscured. 

An early work along these lines investigated the relevance of egocentric object trajectories in the task of next-active-object prediction \cite{furnari2017next}. Provided that an object detector/tracker is available, they proposed to analyze object trajectories observed in a small temporal window to detect next-active-objects before the object-interaction is actually started. 

A more recent work incorporated information on the spatial location of objects into ego-centric action prediction \cite{furnari2019would}. The authors adopted a pretrained deep object detector to localize objects and merged their deep features with another two feature modalities (\ie RGB and optical flow) captured globally from the entire frame. However, their major inference engine is a novel Rolling-Unrolling (RU) LSTM system and object features are simply used as input. In an ablation study, they found that inclusion of the object information helped achieve much better results compared to neglecting that information.

As mentioned above, there are other efforts that treat ego-data no differently from regular third person data.
One such approach processed ego-centric video with a standard deep neural network action recognition model pretrained on third person video datasets \cite{damen2018scaling,wang2015towards}, but now fine-tuned for ego-action future prediction assisted by top-loss \cite{furnari2018leveraging}. Other research made use of standard third person analysis tools to anticipate events that might occur in different places than those currently in view \cite{rhinehart2017first, bokhari2016long}. For example, the current view is of the living room, while the future activity will occur in the kitchen.
In response, they proposed to merge 3D point cloud location information obtained from a SLAM system with interactive objects and high-level scene labels to do the predicting 

Another interesting use of ego-centric data is exploiting the prior knowledge behind the ego-centric data for unsuperivsed learning \cite{bertasius2017using}. The authors researched predicting the next cooperative player with regard to the ego-centric viewer. 
Instead of manually annotating the groundtruth target, they generated pseudo groundtruth in an unsupervised fashion by using ego-centric assumptions; see Figure \ref{fig:gaze} (b).

\section{Datasets and Performance}
In this section, we describe datasets, evaluation metrics and performance in evaluation of the majority of future action prediction approaches discussed above.

\pagebreak
\begin{landscape}
\begin{longtable}{@{}@{} l r r r r r r}
\toprule%
 \centering
 & \multicolumn{1}{c}{{{\bfseries Year}}}
 & \multicolumn{1}{c}{{{\bfseries No. Videos}}}
 & \multicolumn{1}{c}{{{\bfseries No. Actions}}}
 & \multicolumn{1}{c}{{{\bfseries Avg. Len}}}
 & \multicolumn{1}{c}{{{\bfseries Domain}}}
 & \multicolumn{1}{c}{{{\bfseries Data-Modality}}} \\

\cmidrule[0.4pt](l{0.25em}){1-1}%
\cmidrule[0.4pt](r{0.25em}){2-2}%
\cmidrule[0.4pt](r{0.25em}){3-3}%
\cmidrule[0.4pt](r{0.25em}){4-4}%
\cmidrule[0.4pt](r{0.25em}){5-5}%
\cmidrule[0.4pt](r{0.25em}){6-6}%
\cmidrule[0.4pt](r{0.25em}){7-7}%
\endhead

\myrowcolour
\textbf{BDD100K} \cite{yu2020bdd100k} & 2020 &
100,000 & 3 & \highest{40s} & \highest{Car-Stop} & RGB \\ 

\textbf{EGTEA} \cite{fathi2011learning} & 2019 & 
39,596 & 149 & \highest{-} & \highest{Human Interaction} & RGB, HandMask, Gaze \\

\myrowcolour
\textbf{PIE} \cite{rasouli2019pie} & 2019 &
56,000 & 60 & \highest{-} & \highest{Action} & RGB, Skeleton\\

\textbf{Epic-Kitchen} \cite{damen2018scaling} & 2018 &
272 & 125 & \highest{3s} & \highest{Activity, HOI} & RGB \\ 

\myrowcolour
\textbf{MultiTHUMOS} \cite{yeung2018every} & 2017 & 
400 & 65 &  \highest{4.03s} & \highest{Activity} & RGB \\

\textbf{JAAD} \cite{kotseruba2016joint} & 2017 & 346 &
11 & \highest{6.39s} & \highest{Action} & RGB, BB \\

\myrowcolour
\textbf{CAD120} \cite{koppula2013learning} & 2013 & 
120 & 20 & \highest{-} & \highest{Action, HOI} & RGB+D, Skeleton\\

\textbf{50 Salads} \cite{lea2017temporal} & 2012 &
50 & 17 & \highest{-} & \highest{Action-traffic} & RGB, Temporal-Seg \\

\myrowcolour
\textbf{MPII-Cooking}  \cite{rohrbach2012database} & 2012 &
44 & 65 & \highest{600s} & \highest{Activity-cooking} & RGB \\

\textbf{VIRAT} \cite{oh2011large} & 2011 & 
59 & 10 & \highest{219} & \highest{Activity} & RGB+D, Skeleton\\

\myrowcolour
\textbf{UCI-OPPORTUNITY} \cite{roggen2010collecting} & 2010 &
10h & 5 & \highest{-} & \highest{Action, HOI} & RGB, Sensor \\

\textbf{Human-Object-Interaction}  \cite{gupta2009observing} & 2009 & 
10 & 6 & \highest{-} & \highest{Action} & RGB \\
\bottomrule
\caption[Summary of datasets for future action prediction.]{Summary of datasets for future action prediction.} \\
\end{longtable}
\end{landscape}

\textbf{Epic-Kitchen} \cite{damen2018scaling} is a large first-person video dataset, which is captured by 32 subjects in 32 different kitchens. The videos in this dataset contain daily activities of the subjects, i.e., no scripts are provided to instruct the subjects. This makes this dataset very natural and challenging. There are 272 training videos, which are captured by 28 subjects. Each video contains multiple action segments, which are categorized into 125 classes. Since the annotations of the testing videos are not available, the training videos are used to perform cross-validation for evaluation. Specifically, the training videos are randomly split into 7 splits, each containing videos of 4 subjects.

\textbf{50 Salads} \cite{lea2017temporal} contains 50 videos which are performed by 25 subjects. Each subject is preparing two mixed salads. There are 17 fine-grained action classes. Researchers often perform 5-fold cross-validation for evaluation using the splits provided by the authors. There are a total of 50 videos that are filmed over 100 hours.

\textbf{Breakfast} \cite{kuehne2014language} contains 1,712 videos of 52 different actors making breakfast. Overall, there are 48 fine-grained action classes and about 6 action instances for each video. The average duration of the videos is 2.3 minutes and the longest video is 10 minutes.

\textbf{MultiTHUMOS} \cite{yeung2018every} is a challenging dataset for action recognition, containing 400 videos of 65 different actions. On average, there are 10.5 action class labels per video and 1.5 actions per frame and thus it is suitable for long-term activity anticipations.

\textbf{BDD100K} \cite{yu2020bdd100k} consists of 100,000 driving video sequences each 40 seconds of length, accompanied with basic sensory data such as GPS, velocity or acceleration. In total, there are 31k vehicle stopping and 21k not-stopping sequences for training, and 4.6k stopping and 3.1k non-stopping sequences for evaluation.

\textbf{VIRAT} \cite{oh2011large} is a collection of approximately 25 hours of surveillance video taken from various scenes, with an average of 1.6 hours per scene. Multiple HD cameras are used to capture at 1080p or 720p at rates between 25 to 30 Hz. The view angles of cameras towards dominant ground planes ranged between 20 and 50 degrees by stationing cameras mostly at the top of buildings to record a large number of event instances across area, while avoiding occlusion as much as possible.

\textbf{PIE} \cite{rasouli2019pie} comprises 1,842 pedestrian tracks captured using an on-board monocular camera, while driving in urban environments with various street structures and crowd densities. Overall, the ratio of pedestrian non-crossing to crossing events is 2.5 to 1. All video sequences are collected during daylight under clear weather conditions. The videos are continuous, allowing us to observe the pedestrians from the moment they appear in the scene until they go out of the field of view of the camera.

\textbf{MPII-Cooking} \cite{rohrbach2012database} contains 44 instances of cooking activity, which are continuously recorded in a realistic setting. Predictable high level activities are about preparing 14 kinds of dishes, including: making a sandwich, making a pizza, and making an omelet, etc. There are overall 65 different actionlets as building blocks shared among various cooking activities, such as cut, pour, shake, and peel.

\textbf{GTEA Gaze} \cite{fathi2011learning} contains 17 sequences of meal preparation activities performed by 14 different subjects, with the spatial resolution of 640 \& $\times$ 480. It contains the subjects’ gaze location in each frame and the corresponding activity labels. Its extension, \textbf{GTEA Gaze++} \cite{fathi2012learning}, contains 37 sequences performed by 6 subjects of preparing 7 types of meals. Their latest collection, EGTEA \cite{li2018eye}, comes with 10, 321 action instances from 19 verb, 53 noun and 106 action classes.

\textbf{UCI-OPPORTUNITY} \cite{roggen2010collecting} was created in a sensor-rich environment for the machine recognition of human activities. They deployed 72 sensors of 10 modalities in 15 wireless and wired networked sensor systems in the environment, on the objects, and on the human body. The data are acquired from 12 subjects performing morning activities, yielding over 25 hours of sensor data. It contains five high level predictable activities (Relaxing, Coffee time, Early Morning, Cleanup, Sandwich time), 13 low level actionlets (e.g., lock, stir, open, release), and 23 interactive objects (e.g., bread, table, glass).

\textbf{CAD120} \cite{koppula2013learning}  has 120 RGB-D videos of four different subjects performing 10 high-level activities. The data is annotated with object affordance and sub-activity labels and includes groundtruth object categories, tracked object bounding boxes and human skeletons. The set of high-level activities are making cereal, taking medicine, stacking objects, unstacking objects, microwaving food, picking objects, cleaning objects, taking food, arranging objects, having a meal. The set of sub-activity labels are reaching, moving, pouring, eating, drinking, opening, placing, closing, scrubbing, null and the set of affordance labels are reachable, movable, pourable, pourto, containable, drinkable, openable, placeable, closable, scrubbable, stationary.

\textbf{HOI} \cite{gupta2009observing} is the Maryland Human-Object Interactions dataset, which consists of six annotated activities: answering a phone call, making a phone call, drinking water, lighting a flash, pouring water into container and spraying. These activities have about three to five action units each. Constituent action units share similar human movements: 1) reaching for an object of interest, 2) grasping the object, 3) manipulating the object, and 4) putting back the object. For each activity, there are 8 to 10 video samples.

\textbf{Evaluations} for future action prediction can be grouped into two division: \textbf{1.} Reporting the mean average precision of future action label classifications at a single pre-planned future time horizon; \textbf{2.} Reporting the mean average precision over a range of time horizons. The former is widely adopted for activity singleton or sequence anticipations, while the latter for joint activity and time duration anticipations.

In this report, we summarize dataset performances that are frequently  and extensively investigated in recent years regarding the above two divisions. Additional results can be found in the original papers. For the first division, we showcase the next activity anticipation results on the Epic-Kitchen dataset \cite{damen2018scaling}. In its standard setting, the prediction horizon is fixed as the future \textbf{1s}. Epic-Kitchen dataset has complete annotations for fine-grained sub-action denoted as \textbf{Verb}, interactive objects as \textbf{Noun} and high-level action category as \textbf{Action}. Accuracies for all three attributes will be reported. Two sets of accuracy, namely top-1 that only accept the best prediction for evaluation and top-5 that would accept 5 best guesses as final results, are listed to shed light on the robustness of their models; see Table \ref{tab:epic_rst}.

\begin{table}[htbp]
  \centering
  \caption[Future action prediction accuracy on future 1 second for Epic-Kitchen.]{Future action prediction accuracy on future 1 second for Epic-Kitchen. ``set1'' stands for the test set whose scenes appear in the training set, whereas ``set2'' for the test set that does not share scenes with the training set.}
    \begin{tabular}{c|cccc}
    \toprule
    \multicolumn{1}{c}{\multirow{2}[2]{*}{}} & \multirow{2}[2]{*}{Method} & \multicolumn{3}{c}{Top/Top5 Accuracy} \\
    \multicolumn{1}{c}{} &       & Verb  & Noun  & Action \\
    \midrule
    \midrule
    \multirow{6}[1]{*}{set1} & 2S-CNN \cite{damen2018scaling} & 29.76/76.03 & 15.15/38.65 & 4.32/15.21 \\
          & TSN \cite{damen2018scaling}  & 31.81/76.56 & 16.22/42.15 & 6.00/18.21 \\
          & TSN+MCE \cite{miech2019leveraging} & 27.92/73.59 & 16.09/38.32 & 10.76/25.28 \\
          & Trans R(2+1)D \cite{miech2019leveraging} & 30.74/76.21 & 16.47/42.72 & 8.74/25.44 \\
          & RULSTM \cite{furnari2019would} & 33.04/79.55 & 22.78/50.95 & 14.39/33.73 \\
          & FHOI \cite{liu2019forecasting} & 36.25/79.15 & 23.83/51.98 & 15.42/34.29 \\
    \midrule
    \midrule
    \multirow{6}[0]{*}{set2} & 2S-CNN \cite{damen2018scaling} & 25.23/68.66 & 9.97/27.38 & 2.29/9.35 \\
          & TSN  \cite{damen2018scaling} & 25.30/68.32 & 10.41/29.50 & 2.39/9.63 \\
          & TSN+MCE \cite{miech2019leveraging} & 21.27/63.66 & 9.90/25.50 & 5.57/25.28 \\
          & Trans R(2+1)D \cite{miech2019leveraging} & 28.37/69.96 & 12.43/32.20 & 7.24/19.29 \\
          & RULSTM \cite{furnari2019would} & 27.01/69.55 & 15.19/34.28 & 8.16/21.20 \\
          & FHOI \cite{liu2019forecasting} & 29.87/71.77 & 16.80/38.96 & 9.94/23.69 \\
    \end{tabular}%
  \label{tab:epic_rst}%
\end{table}%

For the second division, we follow the Mean-over-Class (MoC) evaluation metrics to measure the performance of joint activity category and duration anticipation on two datasets, Breakfast \cite{kuehne2014language} and 50Salads \cite{lea2017temporal}; see Tables \ref{tab:breakfast_rst} and \ref{tab:breakfast_rst_est} for Breakfast and Tables \ref{tab:50salads_rst} and \ref{tab:50salads_rst_est} for 50Salads. The evaluation method, MoC, represents the average accuracy over the entire prediction horizon.

For singleton activity prediction, as seen in Table \ref{tab:epic_rst}, the current performance is far from satisfactory and thus definitely deserves more effort. For the joint anticipation of activity and time duration, the performance produced from methods that take groundtruth (action, time) observations as input achieves decent results even for remote time indices (\eg 75\% accuracy for 50\% future video horizon predictions for the Breakfast dataset \cite{gammulle2019forecasting}, as seen in Table \ref{tab:50salads_rst}). However, given (action, time) inputs from off-the-self algorithms (Table \ref{tab:breakfast_rst_est} and \ref{tab:50salads_rst_est}), which do not guarantee noisy-free estimation, performance on both datasets drops dramatically. This observation reflects the fact that the current bottle-neck is the accuracy of action detection and segmentation methods more so than the prediction algorithms.

\begin{landscape}
\begin{table}[t]
    \centering 
    \setlength{\tabcolsep}{0.7em} 
    {\renewcommand{\arraystretch}{1.3}
\noindent\begin{tabular*}{\linewidth}{l @{\extracolsep{\fill}} c c c c  c c c c} 
\toprule 
Observation & \multicolumn{4}{c}{20\% } & \multicolumn{4}{c}{30\%} \\ 
\cmidrule(l){2-5}
\cmidrule(l){6-9}
Prediction & 10\% & 20\% & 30\% & 50\% &  10\% & 20\% & 30\% & 50\% \\ 
\midrule 
RNN \cite{abu2018will} & 0.6035 & 0.5044 & 0.4528 & 0.4042 & 0.6145 & 0.5025 & 0.4490 & 0.4175 \\ 
CNN \cite{abu2018will} & 0.5797& 0.4912 & 0.4403 & 0.3926 & 0.6032 & 0.5014 & 0.4518 & 0.4051\\ 
TOS-Dense \cite{ke2019time} & 0.6446 & 0.5627 & 0.5015 & 0.4399 & 0.6595 & 0.5594 & 0.4914 & 0.4423 \\ 
R-HMM (\textbf{Avg})\cite{abu2019uncertainty} & 0.5039 & 0.4171 & 0.3779 & 0.3278 & 0.5125 & 0.4294 & 0.3833 & 0.3307\\
R-HMM (\textbf{Max})\cite{abu2019uncertainty} & 0.7884 & 0.7284 & 0.6629 & 0.6345 & 0.8200 & 0.7283 & 0.6913 & 0.6239 \\
NDR-GAN (\textbf{Avg})\cite{zhaodiverse} & 0.7222 & 0.6240 & 0.5622 & 0.4595 & 0.7414 & 0.7132 & 0.6530 & 0.5238 \\ 
NDR-GAN (\textbf{Max})\cite{zhaodiverse} & 0.8208 & 0.7059 & 0.6851& 0.6406 & 0.8336 & 0.7685 & 0.7213 & 0.6406 \\ 
NeualMemory \cite{gammulle2019forecasting} & 0.8720 & 0.8524 & 0.8102 & 0.7547 & 0.8790 & 0.8279 & 0.8210 & 0.7630 \\ 
\midrule 
\end{tabular*}
}
\caption[Results on Breakfast dataset with groundtruth observations.]{Breakfast Dataset results of dense anticipation mean over classes (MoC) accuracy with groundtruth (action, time) observations as inputs. \text{Avg} stands for averaged results across 16 samplings, while \text{Max} stands for taking the best result among 16 samples.}
\label{tab:breakfast_rst}
\end{table}
\begin{table}[t]
    \centering 
    \setlength{\tabcolsep}{1em} 
    {\renewcommand{\arraystretch}{1.3}
\noindent\begin{tabular*}{\linewidth}{l @{\extracolsep{\fill}} c c c c  c c c c} 
\toprule 
Observation & \multicolumn{4}{c}{20\% } & \multicolumn{4}{c}{30\%} \\ 
\cmidrule(l){2-5}
\cmidrule(l){6-9}
Prediction & 10\% & 20\% & 30\% & 50\% &  10\% & 20\% & 30\% & 50\% \\ 
\midrule 
RNN \cite{abu2018will} & 0.1811 & 0.1720 & 0.1594 & 0.1581 & 0.2164 & 0.2002 & 0.1973 & 0.1921 \\ 
CNN \cite{abu2018will} & 0.1790 & 0.1635 & 0.1537 & 0.1454 & 0.2244 & 0.2012 & 0.1969 & 0.1876\\ 
TOS-Dense \cite{ke2019time} & 0.1841 & 0.1721 & 0.1642 & 0.1584 & 0.2275 & 0.2044 & 0.1964 & 0.1975 \\ 
R-HMM (\textbf{mode})\cite{abu2019uncertainty} & 0.1671 & 0.1540 & 0.1447 & 0.1420 & 0.2073 & 0.1827 & 0.1842 & 0.1686\\
TCN-Cycle \cite{farha2020long} & 0.2588 & 0.2342 & 0.2242 & 0.2154 & 0.2966 & 0.2737 & 0.2558 & 0.2520 \\
Attn-GRU \cite{ng2020forecasting} & 0.2303 & 0.2228 & 0.2200 & 0.2085 & 0.2650 & 0.2500 & 0.2408 & 0.2361 \\
\midrule 
\end{tabular*}
}
\caption[Results on Breakfast dataset without groundtruth observations.]{Breakfast Dataset results of dense anticipation mean over classes (MoC) accuracy \textbf{without} groundtruth (action, time) observations as inputs.}
\label{tab:breakfast_rst_est}
\end{table}
\end{landscape}

\begin{landscape}
\begin{table}
\centering 
    \setlength{\tabcolsep}{1em} 
    {\renewcommand{\arraystretch}{1.3}
\noindent\begin{tabular*}{\linewidth}{l @{\extracolsep{\fill}} c c c c  c c c c} 
\toprule 
Observation & \multicolumn{4}{c}{20\% } & \multicolumn{4}{c}{30\%} \\ 
\cmidrule(l){2-5}
\cmidrule(l){6-9}
Prediction & 10\% & 20\% & 30\% & 50\% &  10\% & 20\% & 30\% & 50\% \\ 
\midrule 
RNN \cite{abu2018will} & 0.4230 & 0.3119 & 0.2522 & 0.1682 & 0.4419 & 0.2951 & 0.1996 & 0.1038 \\ 
CNN \cite{abu2018will} & 0.3608 & 0.2762 & 0.2143 & 0.1548 & 0.3736 & 0.2478 & 0.2078 & 0.1405\\ 
TOS-Dense \cite{ke2019time} & 0.4512 & 0.3323 & 0.2759 & 0.1727 & \textbf{0.4640} & 0.3480 & 0.2524 & 0.1384 \\ 
R-HMM (\textbf{Avg}) \cite{abu2019uncertainty} & 0.3495 & 0.2805 & 0.2408 & 0.1541 & 0.3315 & 0.2465 & 0.1884 & 0.1434\\
R-HMM (\textbf{Max}) \cite{abu2019uncertainty} & 0.7489 & 0.5875 & 0.4607 & 0.3571 & 0.6739 & 0.5237 & 0.4673 & 0.3664 \\
NDR-GAN (\textbf{Avg})\cite{zhaodiverse} & 0.4663 & 0.3562 & 0.3191 & 0.2137 & 0.4613 & 0.3637 & 0.3310 & 0.1945\\ 
NDR-GAN (\textbf{Max})\cite{zhaodiverse} & 0.5150 & 0.3845 & 0.3606 & 0.2762 & 0.5079 & 0.4754 & 0.3783  &  0.2908  \\
NeualMemory \cite{gammulle2019forecasting} & 0.6996 & 0.6433 & 0.6271 & 0.5216 & 0.6810 & 0.6229 & 0.6118 & 0.5667 \\ 
\midrule 
\end{tabular*}
}
\caption[Results on 50Salads dataset with groundtruth observations.]{Results for the 50Salads dataset of dense anticipation mean over classes (MoC) accuracy with groundtruth (action, time) observations as inputs.} 
\label{tab:50salads_rst}
\end{table}

\begin{table}
\centering 
    \setlength{\tabcolsep}{1em} 
    {\renewcommand{\arraystretch}{1.3}
\noindent\begin{tabular*}{\linewidth}{l @{\extracolsep{\fill}} c c c c  c c c c} 
\toprule 
Observation & \multicolumn{4}{c}{20\% } & \multicolumn{4}{c}{30\%} \\ 
\cmidrule(l){2-5}
\cmidrule(l){6-9}
Prediction & 10\% & 20\% & 30\% & 50\% &  10\% & 20\% & 30\% & 50\% \\ 
\midrule 
RNN \cite{abu2018will} & 0.3006 & 0.2543 & 0.1874 & 0.1349 & 0.2164 & 0.2002 & 0.1973 & 0.1921 \\ 
CNN \cite{abu2018will} & 0.2124 & 0.1903 & 0.1598 & 0.0987 & 0.2914 & 0.2014 & 0.1746 & 0.1086\\ 
TOS-Dense \cite{ke2019time} & 0.3251 & 0.2761 & 0.2126 & 0.1599 & 0.33512 & 0.2705 & 0.2205 & 0.1559 \\ 
R-HMM (\textbf{mode}) \cite{abu2019uncertainty} & 0.2486 & 0.2237 & 0.1988 & 0.1282 & 0.2910 & 0.2050 & 0.1528 & 0.1231\\
TCN-Cycle \cite{farha2020long} & 0.3476 & 0.2841 & 0.2182 & 0.1525 & 0.3439 & 0.2370 & 0.1895 & 0.1589 \\ %
Attn-GRU \cite{ng2020forecasting} & 0.3932 & 0.3139 & 0.2701 & 0.2388 & 0.4173 & 0.3273 & 0.3144 & 0.2639 \\
ACC-Grammar \cite{piergiovanni2020adversarial} & 0.3950 & 0.3320 & 0.2590 & 0.2120 & 0.3950 & 0.3150 & 0.2640 & 0.1980 \\
\midrule 
\end{tabular*}
}
\caption[Results on 50Salads dataset without groundtruth observations.]{Results for the 50Salads dataset of dense anticipation mean over classes (MoC) accuracy \textbf{without} groundtruth (action, time) observations as inputs.} 
\label{tab:50salads_rst_est}
\end{table}
\end{landscape}

\chapter{Conclusion and Future Work}\label{CH4}
\section{Current trends}
Throughout this survey, various video predictive tasks have been discussed, among which early action recognition and future action prediction have been studied in a detailed manner. Even though recent efforts have advanced the performance of both tasks greatly, in this section we summarize current status and discuss the potential urgent challenges awaiting to be solved.

\subsection{Early action recognition}
Current trending approaches on early action recognition greatly rely on success from the video action recognition area, where large scale datasets and powerful video discriminative feature representation are frequently advanced via deep learning. As examples: Recent progress comes from use of enhanced video features (\eg \cite{zhao2019spatiotemporal,kong2018adversarial} Temporal segment network \cite{wang2015towards}, \cite{wang2019progressive} ResNet \cite{he2016deep} and \cite{shi2018action} Inception network \cite{szegedy2015going}). Among this progress, the mutual information based (Section 3.2.2) and the propagation based (Section 3.2.3) are the front runners in terms of performance on recent benchmark datasets. These two approaches share the intuition that partial information needs to be augmented, yet they follow different methodologies to achieve that end.

Mutual information based approaches focus on reconstructing a complete video representation through feature matching or knowledge distillation. In these approaches, partial features are mapped to global features, mostly with feature-wise Euclidean distance measuring the feature similarity, with little or no explicit temporal modeling.
In contrast, propagation based approaches focus on the temporal evolution of videos and endeavor to recover useful features at each time step. Some classic temporal sequential models (\eg Kalman filtering and Markov Decision process) are frequently revisited in these approaches; see \cite{zhao2019spatiotemporal, zeng2017visual}.

For early action recognition, a remaining demand is extending study to un-constrained environments. Popular datasets (\eg UCF-101 \cite{soomro2012ucf101}) in existing research are known to be carefully trimmed and controlled.  Though they are collected from real-life video sources and called ``in the wild'', the contained motion is almost always clean, visible and smooth. Such quality is not always guaranteed in many real-life environments. Moreover, the fact that early action recognition still performs poorly in the BIT dataset indicates that most current methods are sensitive to the dataset, \eg between class variation is much greater in UCF-101 compared to BIT.

Even though some research has revealed that certain action categories are relatively easy to recognize early \cite{kong2018adversarial}, it would be favorable to understand the reason behind the observed patterns. For datasets that can be readily recognized early on (\eg UT-Interaction), the decisive factor for the success is still unknown. Also as noted by \cite{zhou2018temporal}, for more complex daily actions that would include compound reasoning structure, \ie ``trying to pour water into a glass, but missing so it spills next to it" from the Something2something dataset, the accuracy is extremely unsatisfactory (around $10\%$ accuracy), indicating that current high accuracy on most examined datasets does not equal to the desired reasoning ability in computational intelligence. 

Another critical issue is the current evaluation procedure for early action recognition. As mentioned in Section 3.1, action videos are often being divided into $k \in (10\%, 100\%)$ progress levels and the recognition accuracy with regard to each level will be reported. However, such a split design may be too coarse to allow for understanding of real actions that unfold more incrementally. For some actions that can be completely executed within 10\% of frames, a high recognition accuracy is misleading. 
Furthermore, little has been done to consider the relative importance of early recognition as a function of action (e.g., elder people falling requires earlier recognition to be useful compared to many other actions).

\subsection{Future Action Prediction}
Mainstream approaches for future action prediction either use temporal sequential tools for building the causal relation between past and long-term future at a high-level (\eg semantics and/or time durations), or make use of visual features (\eg objects, eye gazes, hand motions, poses, etc.) extracted from recently observed frames for anticipating the immediate next action label. In the first research line, the Recurrent Neural Network (RNN) as well as its hybrids with certain traditional sequence models (\eg RNN-HMM \cite{abu2019uncertainty} or Poission-LSTM \cite{mehrasa2019variational}) are the major workhorse, as seen in recent efforts. The high level semantics are often preprocessed with off-the-shelf algorithms and these approaches, in most cases, resorted to a data-driven methodology to learn parameters for both RNNs and classic models. 

In the second direction (\ie more direct use of visual features), the focus is employing multi-modal data with the most effectiveness. Typically, the fusion of distinct information is achieved through one of the following strategies: stacking them together as a single input (\eg \cite{furnari2019would}); designing a multi-task learning framework with extra data modality as regularizers (\eg \cite{liu2019forecasting}); fusing them using a hierarchical structure (\eg \cite{jain2016recurrent, rasouli2020pedestrian}). As means to produce the future action label, researchers often resorted to the recently advanced deep ConvNet features (\eg I3D features \cite{carreira2017quo}) and learned a classifier from the training data.

For joint activity and time duration prediction, it is still an open question as to how to represent times most effectively. Modelling time as a discrete variable seems to perform well, yet it fails when any future time duration exceeds its pre-defined scope. In contrast, the continuous representation would either be too coarse (\eg real value regression) or constrained into a uni-modal distribution (\eg Gaussian), which naturally lacks the ability to model multi-modality of times (\eg \textit{boiling\_egg} can take longer time for a hard boiled egg as well as shorter time for a soft boiled one).

Currently, the prediction of activities often happens in a narrow spatial scope, such as kitchen areas or cooking stations. Yet, daily activities often take place with a transition of locations, \ie fetching objects in another room can incur a scene change.
Even for the ambitious Epic-Kitchen dataset \cite{damen2018scaling}, the default assumption is that all objects relevant for prediction are visible in the current frame. Indeed, there seems to be only one effort that considers prediction across multiple locations \cite{rhinehart2017first} and the dataset has yet to be shared.

As discussed in Section 3.2, in long-term future action predictions, researchers typically resort to the mapping between the high-level semantics \cite{li2014prediction} (\eg model a mapping from a sequence of observed activities to a sequence of future activity labels).
Indeed, in some cases researchers have gone to the extreme of taking groundtruth labels for the observed portion of the video as input.
This setting is reasonable in that an extended video clip contains many frames, which prohibit online low-level feature computation as well as storage, and high-level semantics summarize long videos in a manageable and meaningful way.
Nonetheless, this scheme ignores the rich visual elements that have been demonstrated to be worth exploiting in other areas and which are worth further consideration in long-term predictions.
Moreover, approaches that employ groundtruth labels as input avoid the critical issue of how to extract the labels video in the first place.

\section{Future Work}
In this final section, several directions for future work are suggested.

\textbf{Improved interpretability} of algorithms and representations is necessary to truly understand the early action recognition and future action prediction tasks and their enabling mechanisms. 
From the perspective of understanding the learning methods, there is a demand to further research \textit{what has been learned} by proposed solutions. As an example: In student-teacher learning \cite{wang2019progressive}, it would be beneficial to monitor and understand the teaching process. In the example of deep Kalman feature propagation \cite{zhao2019spatiotemporal}, answers are needed for questions such as how does learning based Kalman filtering affect the error accumulation and what are the motion kernels learned for feature propagation (\eg are they mimicking a warping function)? Both analytic (e.g., \cite{hadji2017spatiotemporal}) and visualization methodologies (e.g., \cite{feichtenhofer2020deep}) have potential for shedding light on these questions.

\textbf{Integrating feature extraction with prediction} is an urgent concern for most high-level based approaches. It has been seen that a decent semantic estimation of input partial observations plays a critical role in making satisfactory long-term prediction (\eg see Table \ref{tab:breakfast_rst} and Table \ref{tab:50salads_rst}), while a noisy estimation deteriorates the performance remarkably (\eg see Table \ref{tab:breakfast_rst_est} and Table \ref{tab:50salads_rst_est}). To tackle this discrepancy, it might be worth integrating the future prediction into the semantic extraction, with an objective to improve the robustness of future prediction with noisy inputs.

\textbf{Better modelling of long-term correlation} is necessary to achieve scalable prediction of activities. The cue that can precisely indicate the future could possibly hide in details that are far away from current observations, and thus it is essential to keep logging historical information and effectively spot the correlations between minor details and future events. Though some efforts have been made for action recognition (\eg the authors in \cite{wu2019long} designed a long-term filter bank to keep a track of remote historical features), its effects on other areas stays unknown and it is likely beneficial to revisit such an ability for future action predictions.

\textbf{Stronger temporal sequence modelling tools} are always favorable to video understanding. The majority of mathematical tools mentioned in Chapter 2 are theoretically grounded and widely adopted. Recently, however, it has been proposed to explore other advances, such as temporal convolution \cite{lea2017temporal}, which challenges the use of memory hidden states, or transformers \cite{girdhar2019video}, which get rid of the recurrent structure and yet demonstrates superior results, as alternatives to classic sequence modelling tools. These innovate findings break the traditional design where the temporal modelling has to be sequential and mostly Markovian. Such directions are worth further investigation. 

\textbf{Learning efficiency} is another concern that closely associates with most recent learning approaches, since these methods prerequisite sufficient size of datasets and expensive annotations, which incurs huge amounts of human labeling efforts, data cleaning and unavoidable data biases. So, how to learn meaningful information without strong supervision or with noisy labels awaits to be answered. Besides, there is also a need for an adaptive learning framework that can assimilate unseen data and new action categories, since people's daily activities would evolve along with their own development. An active and continuous learning framework would be of great benefit.
Unsupervised approaches to representational learning, e.g., as introduced under the topic of predictive video coding in Chapter 1, provide a potentially useful avenue to explore to address these concerns.

\textbf{Creation of better datasets and experimental protocols} is necessary to better validate prediction algorithms. From the perspective of datasets and evaluation metrics, it is worth understanding the specific observation ratios that are practically valuable in real-life conditions with regard to actions, and therefore re-evaluating existing approaches using this new criterion. It is also worth investigating the correlation between the video observation ratio and action evolution ratio of popular datasets, in order to examine the validation of current early recognition evaluation metrics. 

\section*{Acknowledgements}
The authors thank Michael S. Brown and Kosta G. Derpanis for the insightful comments they provided on this review.


\ssp
\renewcommand{\bibname}{References}
\bibliographystyle{plain}
\bibliography{APPENDIX/References}

\end{document}